\documentclass[11pt]{article}

\usepackage[final]{acl}

\usepackage{times}
\usepackage{latexsym}

\usepackage[T1]{fontenc}

\usepackage[utf8]{inputenc}

\usepackage{textcomp}

\usepackage{amssymb}

\usepackage{microtype}

\usepackage{inconsolata}

\usepackage{graphicx}

\usepackage{booktabs} 
\usepackage{ulem} 
\usepackage{multirow}
\usepackage{amsmath}
\usepackage{caption}
\usepackage{cuted}
\usepackage{capt-of}
\usepackage{colortbl}
\usepackage[most]{tcolorbox}

\usepackage{listings}
\usepackage{xcolor}

\lstdefinestyle{mystyle}{
	backgroundcolor=\color{gray!5},
	basicstyle=\ttfamily\footnotesize,
	breaklines=true,
	frame=single,
	rulecolor=\color{gray!60},
	keywordstyle=\color{blue},
	commentstyle=\color{gray!70}\itshape,
	stringstyle=\color{teal},
	showstringspaces=false,
}
\tcbset{colback=white, colframe=black!30, boxrule=0.5pt, arc=2mm}

\newcommand{\method}{MAViS}

\title{\method: A \underline{M}ulti-\underline{A}gent Framework for Long-Sequence \underline{Vi}deo \underline{S}torytelling}

\author{Qian Wang \\
	Virginia Tech \\
	USA \\
	\texttt{yqwq1996@vt.edu} \\\And
	Ziqi Huang \\
	Nanyang Technological University \\
	Singapore \\
	\texttt{ZIQI002@e.ntu.edu.sg} \\\And
	Ruoxi Jia \\
	Virginia Tech \\
	USA \\
	\texttt{ruoxijia@vt.edu} \\\AND
	Paul Debevec \\
	Eyeline Studios \\
	USA \\
	\texttt{debevec@scanlinevfx.com} \\\And
	Ning Yu \\
	Eyeline Studios \\
	USA \\
	\texttt{ning.yu@scanlinevfx.com} \\}

\begin{document}
	\maketitle
	
	\begin{strip}
		\centering
		\includegraphics[width=\textwidth]{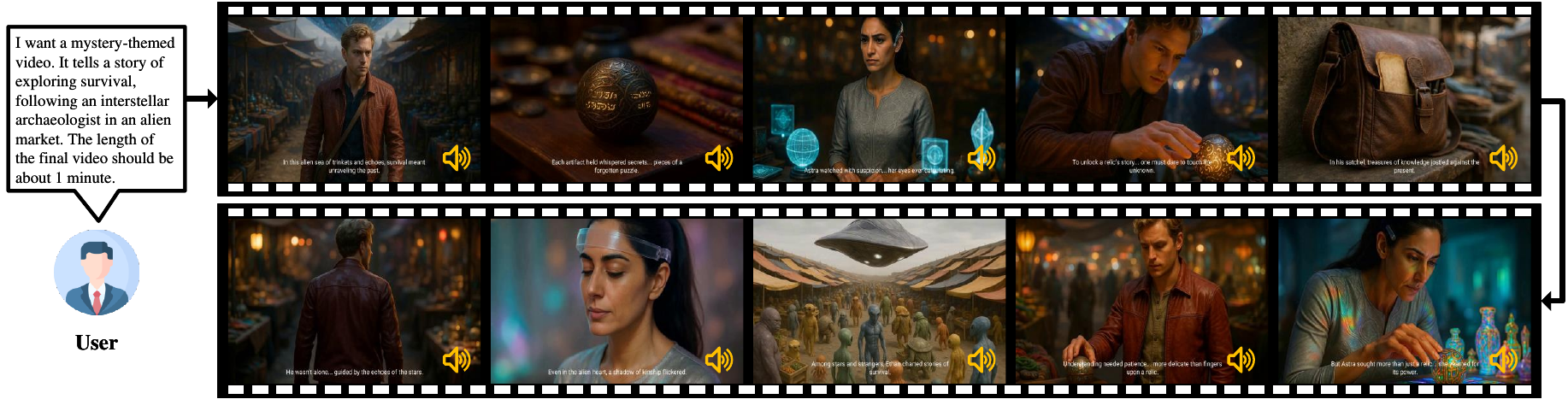}
		\captionof{figure}{
			With just a brief prompt, \method \ enables users to rapidly explore diverse visual storytelling and creative directions for sequential video generation by efficiently producing high-quality, complete long-sequence videos.}
		\label{fig:teaser}
		\vspace{-2mm}
	\end{strip}

	\begin{abstract}
		Despite recent advances, long-sequence video generation frameworks still suffer from significant limitations: poor assistive capability, suboptimal visual quality, and limited expressiveness. 
		To mitigate these limitations, we propose \textbf{\method}, an end-to-end multi-agent collaborative framework for long-sequence video storytelling. 
		\method \ orchestrates specialized agents across multiple stages, including script writing, shot designing, character modeling, keyframe generation, video animation, and audio generation.
		In each stage, agents operate under the \textbf{3E Principle}—Explore, Examine, and Enhance—to ensure the completeness of intermediate outputs. 
		Considering the capability limitations of current generative models, we propose the Script Writing Guidelines to optimize compatibility between scripts and generative tools. 
		Experimental results demonstrate that \method \ achieves state-of-the-art performance in assistive capability, visual quality, and video expressiveness. 
		Its modular framework further enables scalability with diverse generative models and tools. 
		With just a brief prompt, \method \ enables users to rapidly explore diverse visual storytelling and creative directions for sequential video generation by efficiently producing high-quality, complete long-sequence videos.
		To the best of our knowledge, \method~is the only framework that provides multimodal design output—videos with narratives and background music.

	\end{abstract}

	\section{Introduction}
	
	\begin{table*}[ht]
		\scriptsize
		\centering
		\tabcolsep=0.24cm
		\begin{tabular}{l|cccccccc}
			\toprule
			\textbf{Methods} & \textbf{Agentic} & \textbf{Script Writing} & \textbf{Shot Designing}  & \textbf{Dynamic} & \textbf{ID Consistency} & \textbf{Audio} & \textbf{Automated}  \\
			\midrule
			Mora~\cite{Mora} & \checkmark & $\times$ & $\times$ & \checkmark & $\times$ & $\times$ & \checkmark  \\ 
			AesopAgent~\cite{AesopAgent} & \checkmark & \checkmark & $\times$ & $\times$ & \checkmark & $\times$ & \checkmark  \\
			MovieDreamer~\cite{MovieDreamer} & $\times$ & $\times$ & $\times$ & \checkmark & \checkmark & $\times$ & $\times$\\ 
			DreamFactory~\cite{Dreamfactory} & \checkmark & $\times$ & \checkmark & \checkmark & \checkmark & $\times$ & $\times$ \\ 
			VGoT~\cite{VideoGen-of-Thought} & $\times$ & \checkmark & $\times$ & \checkmark & \checkmark & $\times$ & \checkmark\\ 
			LCT~\cite{LCT} & $\times$ & $\times$ & $\times$ & \checkmark & \checkmark & $\times$ & $\times$ & \\ 
			MovieAgent~\cite{MovieAgent} & \checkmark & $\times$ & \checkmark & \checkmark & \checkmark & \checkmark & $\times$  \\
			\midrule
			\textbf{\method~(ours)} & \checkmark & \checkmark & \checkmark & \checkmark & \checkmark & \checkmark & \checkmark \\
			\bottomrule
		\end{tabular}
		\caption{
			\textbf{Properties of \method \ vs Other Methods}: We summarize seven metrics for long-sequence video storytelling works.
			"Agentic": whether the system employs a multi-agent architecture; "Script Writing": ability generate scripts from user prompt; "Shot Designing.": support for structured shot designing and controllable narrative flow; "Dynamic": whether the generated video exhibits dynamic features; "ID Consistency": consistency in character identity across multiple shots; "Audio": whether the output includes synchronized audio; "Automated": whether the pipeline can operate without the presence of user input in script writing, script planning, or model fine-tuning. 
		} 
		\label{tab: baseline}
	\end{table*}
	
	In recent years, video generative models~\cite{Sora,hailuo,Mochi} have made substantial strides in text-to-video (T2V) and image-to-video (I2V) generation. While these methods demonstrate promising capabilities for shot-level creation, the generation of long videos remains a formidable challenge. 
	Besides, the current scale of computing and data is insufficient to train a single end-to-end model capable of producing high-quality, minute-long videos.

	To overcome this limitation, prior works have explored extending existing video clips~\cite{Kling, Streamingt2v, TATS-longv}, or composing sequences of keyframes~\cite{StoryGen, SEED-story, Make-a-story, AesopAgent}, or integrating storytelling and video generation~\cite{Nuwa-xl, VideoGen-of-Thought}. 
	However, these approaches suffer from repetitive motion, lack of narrative cohesion, or result in static image series with limited visual expressions. 
	To better align narrative planning with video animation, MovieDreamer~\cite{MovieDreamer} adopts an auto-regressive approach to generate key frames from input scripts, while Dreamfactory~\cite{Dreamfactory} and MovieAgent~\cite{MovieAgent} propose a hierarchical script–scene–shot pipeline. 
	However, as shown in Table~\ref{tab: baseline}, these methods require substantial manual effort from users for script writing and training LoRA models to maintain ID consistency, which significantly limits their assistive capability and hinders large-scale deployment.

	In summary, current long-sequence video storytelling methods face three major limitations: 
	1) poor assistive capability, not supporting script writing, shot designing, and LoRA training to maintain ID consistency; 
	2) suboptimal visual quality, manifested in visual distortions, unnatural actions, and inaccurate object proportions; 
	and 3) limited expressiveness, often resulting in repetitive actions and poorly structured shot compositions.

	
	To address this, we introduce \textbf{\method}, a multi-agent framework for end-to-end long-sequence video generation. 
	\method~encompasses various stages, including script writing, shot designing, character modeling, keyframe generation, video animation, and audio generation. 
	Given a brief user prompt, the script writing stage first generates a structured script with character definitions and shot outlines. 
	The shot designing stage defines a series of shot-level elements to guide subsequent content generation, while character modeling ensures ID consistency via LoRA-trained appearances.
	During the keyframe generation stage, T2I models are employed to create the initial static frames for each shot, i.e., the keyframes, which are then expanded into full video sequences in the video animation stage through I2V models, maintaining visual coherence and narrative fluency. 
	Finally, voice-overs and background music (BGM) are added to each shot during the voice generation stage, resulting in a cohesive, long-sequence video.

	Crucially, all stages operate under the \textbf{3E Principle}:  
	- \textit{Explore}: Initial generation of candidate outputs;  
	- \textit{Examine}: Reviewing the quality and completeness of the output;  
	- \textit{Enhance}: Iterative refinement and optimization.  
	After multiple iterations, each stage passes the most complete version of its output to the next, enabling gradual optimization throughout the end-to-end pipeline.
	
	To mitigate the limitations of current generative models in maintaining background consistency, executing complex actions, and rendering fine-grained details, we propose the Script Writing Guidelines aimed at enhancing the compatibility between scripts and generative models, ultimately improving the expressiveness of the video storytelling.
	As illustrated in Figure~\ref{fig:teaser}, with just a brief prompt, \method \ enables users to rapidly explore diverse visual storytelling and creative directions for sequential video generation by efficiently producing high-quality, complete long-sequence videos.
	Furthermore, its modular architecture ensures scalability with various generative tools and models.
	
	
	
	Our contributions are summarized in three thrusts:
	
	$\bullet$ We propose \textbf{\method}, an end-to-end multi-agent framework that encompasses various stages including script writing, shot designing, character modeling, keyframe generation, video animation, and audio generation — forming a complete and scalable workflow for long video storytelling.
	
	$\bullet$ We introduce the \textbf{3E Principle}—\textit{Explore}, \textit{Examine}, \textit{Enhance}—to continuously optimize the output of each stage.
	This iterative design ensures the completeness of the output at each stage.
	
	$\bullet$ We propose the Script Writing Guidelines for long-sequence video storytelling to enhance the script-tool compatibility, ultimately improving the expressiveness of the video storytelling.
	
	$\bullet$ Experiments demonstrate that \method \ achieves state-of-the-art performance and receive the most user preference votes in terms of expressiveness and visual quality. 
	With just a brief user prompt, \method \ efficiently generates high-quality and complete long-sequence video storytelling. 
	To the best of our knowledge, \method~is the only framework that provides multimodal design output.

	\section{Related Works}

	\begin{figure*}[ht]
		\centering
		\includegraphics[width=2.05\columnwidth]{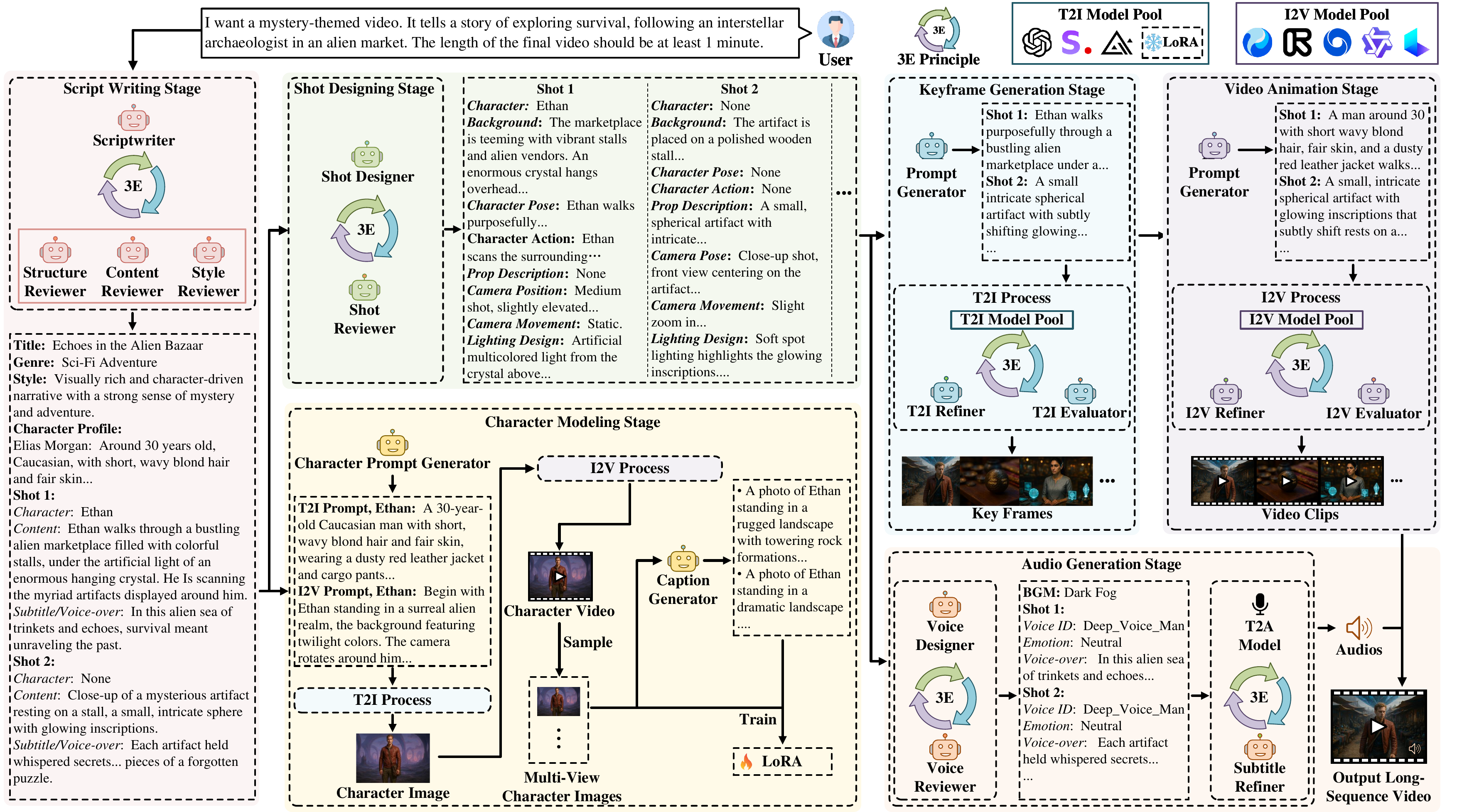}
		\caption{
			Illustration of \method  ~framework.
		}
		\label{fig: pipeline}
	\end{figure*}

	\subsection{Video Generation}
	Long video generation~\cite{LVDM,TATS-longv,wang2024leo} aims to produce longer videos than previous text-to-video (T2V)~\cite{modelscope,Lavie} and image-to-video (I2V)~\cite{EmuVideo,Dynamicrafter,I2VGen} models.
	Auto-regressive approaches~\cite{ART-V, Streamingt2v} and video storytelling methods~\cite{Nuwa-xl, VideoGen-of-Thought} suffer from repetitive motions with limited narrative development and poor visual quality.
	Meanwhile, commercial models~\cite{Kling, Gen3, Gen4} support temporal extension. 
	However, the duration of the generated videos remains constrained.
	In this work, \method \ uses cutting-edge video generation models to synthesize high-quality video clips and assembles them into long-sequence video storytelling. 
	

	\subsection{AI Agents}
	AI agents~\cite{autogpt, Lumos, Camel, Metagpt, GenerativeAgents}, empowered by large language models~\cite{GPT,Llama,Llama2}, have shown broad applicability in reasoning, planning, web interaction, and visual generation. 
	In video generation, systems~\cite{WorldGPT, Mora, AesopAgent} explore world modeling, multi-agent coordination, and retrieval-augmented synthesis, yet often yield static content with limited transitions. 
	Hierarchical pipelines~\cite{Dreamfactory, MovieAgent} introduce script-to-shot workflows but still rely on manual scripting and LoRA tuning. 
	In contrast, our \method \ leverages multi-agent collaboration to enable long-sequence video storytelling.

	\section{\method}
	We propose \textbf{\method}, a multi-agent end-to-end framework for long-sequence video storytelling that consists of various stages, including script writing, shot designing, character modeling, keyframe generation, video animation, and audio generation.
	
	Given a user prompt $P$, as illustrated in Figure~\ref{fig: pipeline} \method \ aims to generate a long-sequence video storytelling that comprises $N$ shots:
	\begin{equation}
		\small
		\mathcal{F}: P \rightarrow \hat{\mathcal{V}}, \quad \hat{\mathcal{V}} = \{\{{V}_j, \ {A}_j\} \mid j = 1, 2, \dots, N\},
	\end{equation}
	where $V_j$ is the video clip and $A_j$ is the audio track.
	To ensure the completeness and reliability of the generation pipeline, agents in each stage follow a unified \textbf{3E Principle}: \textit{Explore}, \textit{Examine}, and \textit{Enhance}. 
	Section~\ref{sec: 3E} details the 3E principle, and Section~\ref{sec: script writing guide} outlines the Script Writing Guidelines, followed by Section~\ref{sec: script writing} to~\ref{sec: production} introducing the agent design and working procedures of each stage.

	\subsection{3E Principle}
	\label{sec: 3E}
	Existing frameworks for long-sequence video storytelling adopt one-shot generation strategies. 
	However, such approaches often prove inadequate in practice—scripts may not align with user intent, images or videos can exhibit distortions or unnatural actions, and audio tracks may be misaligned in duration.
	These issues highlight the necessity of adopting a multi-round generation–modification loop, rather than relying on one-shot generation.
	To address this, we introduce the \textbf{3E Principle}—Explore, Examine, Enhance—to guide intra-stage agents' collaboration and progressively improve generation quality across iterations:
	
	\textit{Explore}: This stage represents the first attempt to generate content based on the current request and specification:  
	\begin{equation}
		\small
		r_1 = G(p_1; g),
	\end{equation}
	where $p_1$ denotes the initial generation request, $g$ is the generation guidelines, such as guides of script-tool compatibility and output completeness. $G$ is the generator, and $r_1$ is the resulting output.
	
	\textit{Examine}: The generated output is then examined to determine whether it satisfies the specification:  
	\begin{equation}
		\small
		f_1 = E(r_1, g),
	\end{equation}
	where $E$ is the reviewer, and $f_1$ is the examination feedback of result $r_1$ based on the guidelines $g$.
	
	\textit{Enhance}: Based on the feedback and guidelines, the request is refined to improve quality in the next iteration:   
	\begin{equation}
		\small
		p_{i+1} = R(p_i; f_i, g),
	\end{equation}
	where $R$ denotes the refiner, and $p_{i+1}$ is the updated request for iteration $i+1$.
	This new request serves as the input to the next round of Explore-Examine-Enhance, and the process continues iteratively until the output fully satisfies the specification.
	
	After multiple Explore-Examine-Enhance iterations, each stage passes the most complete version of its output to the next, enabling gradual optimization throughout the pipeline.
	Given the limitations of individual agents, it may need to deploy multiple reviewers in parallel to ensure comprehensive assessment. 
	Additionally, generating multiple candidates per iteration and enhancing the highest-scoring one can effectively reduce the total number of required iterations.
	Moreover, depending on the design, the generator and refiner may be implemented as a unified component.

	\subsection{Script Writing Guidelines}
	\label{sec: script writing guide}
	Given the limitations of current generative models in maintaining background consistency, executing complex actions, and rendering fine-grained details, certain scripts remain infeasible for video generation. 
	To address this gap and improve the compatibility between scripts and generative models, we propose the Script Writing Guidelines for long-sequence storytelling. 
	The Script Writing Guidelines comprises three components: the Structure Guide, which focuses on how scenes and shots are segmented and arranged; 
	the Content Guide, which defines how much content a single shot should include; and the Style Guide, which ensures user-centric alignment, coherent pacing, logical flow, and structural completeness. 
	Together, these guidelines constrain agent-based script writing and improve the expressiveness of the ultimate video storytelling.
	
	\paragraph{\textbf{Structure Guide}}
	Current video generation models have limited contextual modeling capacity; as a result, adjacent shots set within the same spatial environment often suffer from inconsistency in background elements. 
	Therefore, scriptwriters should avoid placing successive shots in the same background and should minimize transitions between tightly connected spatial locations (e.g., from "outside the door" to "inside the room," or from one room to another). 
	Moreover, to reduce the jarring effect of spatial jumps, it is recommended not to place the same character in consecutive shots unless narrative necessity dictates it.
	
	To mitigate these issues while preserving narrative flow, transitional shots can be inserted. These may include close-ups of different characters, partial views of a character (e.g., hands, back), detailed shots of the environment, or important props. Such insertions help avoid background duplication while reinforcing the spatial or character continuity. 
	Additionally, enhanced descriptions of both background and character appearance are encouraged to improve model alignment with visual and narrative consistency.
	
	\paragraph{\textbf{Content Guide}}
	Current video generation models struggle to synthesize multiple sequential actions within a single shot (e.g., running followed by walking and then dancing). 
	Therefore, it is advisable to limit each shot to a single, simple action. 
	Furthermore, generative models often fail to produce fine-grained visual elements accurately, such as textual content on smartphone screens or small patterns on objects. Scripts should avoid specifying such details.

	\paragraph{\textbf{Style Guide}}
	Within the Structure Guide and the Content Guide, a script may risk becoming fragmented or visually monotonous if not properly orchestrated.
	The style guide thus ensures harmonious pacing, visual diversity, and narrative logic.
	Moreover, the script's tone and theme must align with the user's request, and different genres require distinct narrative pacing—emotional and family-oriented scripts benefit from a slow, layered progression, while action-oriented scripts demand rapid and dynamic developments. 
	Additionally, structural completeness is essential: every script must contain a title, character definitions, and a shot-by-shot script. Each shot should specify the character (or "None" if absent), shot content, and accompanying subtitle or voice-over. Subtitles must align semantically with the shot content.

	\subsection{Script Writing Stage}
	\label{sec: script writing}
	Upon receiving a brief user prompt, four agents—the Scriptwriter, the Structure Reviewer, the Content Reviewer, and the Style Reviewer—work collaboratively in the script writing stage to produce a structured script.
	
	The Scriptwriter $G_{\textrm{scr}}$ first generates a draft script $S_1$ based on the user prompt $P$ and the Script Writing Guidelines $g_{\textrm{scr}}$:
	\begin{equation}
		\small
		S_1 = G_{\textrm{scr}}(P, g_{\textrm{scr}}).
	\end{equation}
	
	Following the 3E principle, at each iteration, the Structure Reviewer $E_{\textrm{str}}$, Content Reviewer $E_{\textrm{con}}$, and Style Reviewer $E_{\textrm{sty}}$ examine the script $S_i$ against the Structure Guide $g_{\textrm{str}}$, Content Guide $g_{\textrm{con}}$ and Style Guide $g_{\textrm{sty}}$ respectively, and provide feedback:
	{\small
		\begin{equation}
			\small
			f_{\mathrm{scr}}^i = \left[ E_{\textrm{str}}(S_i, g_{\textrm{str}}),\ E_{\textrm{con}}(S_i, g_{\textrm{con}}),\ E_{\textrm{sty}}(S_i, g_{\textrm{sty}}) \right], 
		\end{equation}
	}
	where $g_{\textrm{scr}} = [g_{\textrm{str}},\ g_{\textrm{con}},\ g_{\textrm{sty}}]$.
	The Scriptwriter integrates these feedback to improve the script iteratively:
	\begin{equation}
		\small
		S_{i+1} = G_{\textrm{scr}}(f_{\mathrm{scr}}^i;\ S_i, g_{\textrm{scr}})
	\end{equation}

	The Explore-Examine-Enhance loop continues until all reviewers reach a consensus that the script fully satisfies the guidelines after $n$ iterations, and the Scriptwriter outputs the finalized version, represented as $S = C_{\textrm{scr}}(S_n)$.

	\subsection{Shot Designing Stage}
	In the shot-designing stage, the Shot Designer—responsible for generating detailed shot designs—and the Shot Reviewer—tasked with assessing the completeness and adherence of shot designs to predefined guidelines—play pivotal roles in transforming the high-level script into detailed shot elements suitable for downstream image and video generation.
	In detail, each shot is enriched with seven essential elements to enable precise visual control:
	
	$\bullet$ \textit{Background}: Specifies the spatial and environmental context of the shot, including location, time of day, season, weather conditions, and foreground-background composition.
	
	$\bullet$ \textit{Character Pose}: Describes the initial static state of characters, including posture, expression, and position when the shot begins.
	
	$\bullet$ \textit{Character Action}: Captures dynamic interactions or movements of characters throughout the shot.
	
	$\bullet$ \textit{Prop Description}: Defines non-character objects, including their form, texture, behavior, and placement.
	
	$\bullet$ \textit{Camera Position}: Specifies the camera’s initial angle, orientation, and distance from the scene.
	
	$\bullet$ \textit{Camera Movement}: Describes the movement mode and direction of the camera during the shot.
	
	$\bullet$ \textit{Lighting Design}: Defines lighting characteristics such as type, tone, brightness, and color temperature to convey atmosphere.
	
	Similarly, following the 3E Principle, the Shot Designer generates an initial shot design from the script:
	\begin{equation}
		\small
		\hat{S}_1 = G_{\mathrm{shot}}(S),
	\end{equation}
	where the shot designing guide provides essential constraints to ensure completeness and stylistic alignment with the original script.
	
	In each iteration $i$, the Shot Reviewer examines the completeness and appropriateness of the current design $\hat{S}_i$ and provides feedback:
	\begin{equation}
		\small
		f_{\mathrm{shot}}^i = E_{\mathrm{shot}}(\hat{S}_i).
	\end{equation}
	The Shot Designer then uses this feedback to enhance the design, until the Shot Reviewer deems the result satisfactory:
	\begin{equation}
		\small
		\hat{S}_{i+1} = G_{\mathrm{shot}}(f_{\mathrm{shot}}^i;\ \hat{S}_i), \quad \hat{S} = \hat{S}_n.
	\end{equation}

	\subsection{Character Modeling Stage}
	\label{Sec: character stage}
	To ensure the ID consistency of characters across shots, in the character modeling stage, agents construct visual representations for each character and train a LoRA model. 
	This process incorporates the Character Prompt Generator and Caption Generator, supported by reviewer agents for both image generation (T2I Evaluator and T2I Refiner) and video animation (I2V Evaluator and I2V Refiner).
	
	The process begins with the Character Prompt Generator, which generates both text-to-image (T2I) and image-to-video (I2V) prompts for each character. 
	T2I prompts aim to generate a frontal and naturally posed character image, while I2V prompts are used—together with the T2I images—to synthesize character-centric video clips that capture the subject from multiple angles.
	After both T2I and I2V processes, multi-view character images are sampled from the generated character video. 
	The Caption Generator then produces structured captions for each sampled frame. 
	These images and captions are subsequently used to train LoRA models (if needed).
	
	Both the T2I and I2V processes follow the 3E Principle, but differ from script writing and shot designing in that each iteration produces multiple samples from the generative model pools. 
	Specifically, in the $i$-th iteration, a set of candidate images is generated by the pool of T2I models $\mathcal{M}_{I}$ using the current T2I prompt $p_{I}^i$:
	\begin{equation}
		\small
		\mathcal{I}_i = \{ G_{I}^k(p_{I}^i), \ G_{I}^k \in \mathcal{M}_{I}, \ k = 1,2,\dots, |\mathcal{M}_{I}| \}.
	\end{equation}
	T2I Evaluator $E_{I}$ scores candidates according to the image evaluation guide $g_{I}$, selects the best one $I_i$, and provides evaluation result $f^i_I$:
	\begin{equation}
		\small
		[I_i, f^i_I] = E_{I}(\mathcal{I}_i, g_{I})
	\end{equation}
	The T2I Refiner $R_{I}$ then updates the prompt accordingly:
	\begin{equation}
		\small
		p_{I}^{i+1} = R_{I}(f^i_I, p_{I}^i)
	\end{equation}
	After $n$ iterations, the final image $I = I_n$ initiates the I2V process.

	The I2V process mirrors the T2I process. 
	First, a set of candidate videos is generated by the I2V model pool $\mathcal{M}_{V}$:
	\begin{equation}
		\small
		\mathcal{V}_i = \{ G_{V}^k(I, p_{V}^i), \ G_{V}^k \in \mathcal{M}_{V}, \ k = 1,2,\dots, |\mathcal{M}_{V}| \}. 
	\end{equation}
	The I2V Evaluator $E_{V}$ evaluates these candidates according to the video evaluation guide $g_{I}$, selects the best one $V_i$, and provides evaluation result $f^i_V$. 
	The I2V Refiner $R_{V}$ updates the I2V prompt and produce the final video $V_n$ after $n$ iterations:
	\begin{equation}
		\small
		[V_i, f^i_V] = E_{V}(\mathcal{V}_i, g_{V}), \quad  p_{V}^{i+1} = R_{V}(f^i_V, p_{I}^i), \quad V = V_n.
	\end{equation}
	
	\begin{table*}[ht]
		\scriptsize
		\centering
		\tabcolsep=0.216cm
			\begin{tabular}{l|cc|cccccc}
				\toprule
				\multirow{2}{*}{Method} & \multicolumn{2}{c}{Keyframe Generation} & \multicolumn{6}{|c}{Video Animation} \\
				\cline{2-9}
				& CLIP $\uparrow$ & Inception $\uparrow$ & T. Flick. $\uparrow$ & M. Smooth. $\uparrow$ & Sub. Cons. $\uparrow$ & Bg. Cons. $\uparrow$ & Aesthetic $\uparrow$ & I. Quality $\uparrow$\\
				\hline
				VGoT~\cite{VideoGen-of-Thought} & 22.37 & 8.53 & 99.00 & 99.31 & \textbf{98.94} & \textbf{98.74} & \textbf{80.11} & 64.59\\
				
				Mora~\cite{Mora} & 33.98 & 12.68 & 97.32 & 99.23 & 95.23 & 94.88 & 64.36 & 71.48\\
				
				MovieAgent~\cite{MovieAgent} & 31.71 & 10.72 & 97.07 & 99.20  & 93.29 & 94.51 & 63.39 & 71.26  \\
				
				\hline
				\textbf{\method~(ours)} & \textbf{34.22} & \textbf{12.81} & \textbf{99.09} & \textbf{99.53} & 95.72 & 96.12 & 63.17 & \textbf{72.91}\\
				\bottomrule
			\end{tabular}
			\caption{
				\textbf{Evaluation of Automatic Metrics for Keyframe Generation and Video Animation.} 
				"T. Flick.", "M. Smooth.", "Sub. Cons.", "Bg. Cons.", "Aesthetic", and "I. Quality" refer to "Temporal Flickering", "Motion Smoothness", "Subject Consistency", "Background Consistency", ‘Aesthetic Quality’, and "Imaging Quality" from metrics in VBench~\cite{Vbench}, respectively. $\uparrow$ indicates a higher value is more desired. \textbf{Bold} indicates the best results. 
			}
			\label{Tab: automatic}
		\end{table*}

		\begin{table*}[ht]
			\scriptsize
			\centering
			\tabcolsep=0.28cm
			\begin{tabular}{l|cccccccccc}
				\toprule
				Method & Narrative $\uparrow$ & Visual $\uparrow$ & User Align. $\uparrow$ & Sub. Cons. $\uparrow$ & Sub. Natural. $\uparrow$ & Bg. Cons. $\uparrow$ & Bg. Real. $\uparrow$\\
				\hline
				VGoT~\cite{VideoGen-of-Thought} & 3.39 & 3.39 & 6.79 & 5.56 & 3.07 & 4.14 & 3.24 \\
				
				Mora~\cite{Mora} & 15.18 & 16.96 & 13.57 & 15.77 & 13.00 & 13.69 & 15.29 \\
				
				MovieAgent~\cite{MovieAgent} & 9.46 & 11.79 & 11.96 & 12.37 & 12.27 & 12.07 & 10.97  \\
				
				\hline
				\textbf{\method~(ours)} & \textbf{71.96} & \textbf{67.86} & \textbf{67.68} & \textbf{66.31} & \textbf{71.66} & \textbf{70.09} & \textbf{70.50} \\
				\bottomrule
			\end{tabular}
			\caption{
				\textbf{User Study on the Performance of Long-Sequence Video Storytelling} (Voting Results). 
				"Narrative", "Visual", "User Align.", "Sub. Cons.", "Sub. Natural.", "Bg. Cons.", and "Bg. Real." refer to "Narrative Expressiveness", "Visual Quality", "User Prompt Alignment", "Subject Consistency", "Character Naturalness", "Background Consistency",  and "Background Realism", respectively. $\uparrow$ indicates a higher value is more desired. \textbf{Bold} indicates the best results. 
			}
			\label{Tab: user study}
		\end{table*}

		The difference is that the I2V process incorporates additional evaluation complexity.
		The image evaluation guide assesses candidates based on three primary axes: 1) \textit{Visual Quality}: clarity, detail resolution, and aesthetic harmony; 2) \textit{Naturalness}: identifying distortions, anatomical inaccuracies, or proportion inconsistencies; and 3) \textit{Prompt Consistency}: if the image content aligns with the input prompt.
		In contrast, the video evaluation guide extends these criteria by introducing: 1) \textit{Subject Consistency}: temporal stability in facial features, attire, hairstyle, and skin tone across frames; and 2) \textit{Dynamics}: the richness, fluidity, and pacing of character movement throughout the clip.
		Together, the image and video evaluation guides work in concert to enforce the completeness of the final generation.

		\subsection{Generation Stages}
		\label{sec: production}
		The generation stages include: 1) the keyframe generation stage, which comprises the T2I Prompt Generator, T2I Evaluator, and T2I Refiner, responsible for generating the initial reference image for each shot; 2) the video animation stage, which includes the I2V Prompt Generator, I2V Evaluator, and I2V Refiner, and completes the video shot based on the keyframes; and 3) the audio generation stage, which consists of the Voice Designer, Voice Reviewer, and Subtitle Refiner, synthesizes voice-overs and background music aligned with the emotional rhythm of each shot.
		
		First, in the keyframe generation stage, the T2I Prompt Generator uses shot-level elements—\textit{Character}, \textit{Background}, \textit{Character Pose}, \textit{Prop Description}, \textit{Camera Position}, and \textit{Lighting Design}—to construct T2I prompts, which are refined via the same T2I process in Section~\ref{Sec: character stage}, utilizing the T2I model pool and the trained LoRA model, resulting in a set of keyframes $\{I_j \mid j = 1, 2, \dots, N\}$.

		Second, in the video animation stage, the I2V Prompt Generator formulates I2V prompts using \textit{Character}, \textit{Background}, \textit{Character Action}, \textit{Prop Description}, \textit{Camera Movement}, and \textit{Lighting Design}. 
		Again, following the same I2V process introduced in Section~\ref{Sec: character stage}, these prompts are used to generate video clips $\{V_j \mid j = 1, 2, \dots, N\}$.
		
		After video animation, voice and music planning are initiated in the audio generation stage based on the 3E principle.
		The Voice Designer $G_A$ selects background music for the entire script and configures voice IDs and emotional tones for each individual shot. 
		The voice design is then reviewed by the Voice Reviewer $E_A$, who checks for compliance (e.g., music availability and voice–emotion appropriateness) and provides iterative feedback $f_A^i$ to assist in refinement:
		{
			\small
			\begin{align}
				\label{eq:space_reserve}
				S_A^{1} = G_A(S), \quad& f_A^i = E_A(S_A^i), \quad  S_A^{i+1} = G_A(S, f_A^i),\nonumber\\
				& \quad S_A = S_A^n.
			\end{align}
		}

		Next, the T2A model $G_A$ generates speech audio for each shot. 
		The Subtitle Refiner $R_A$ shortens the voice-over $p_A$ to fit temporal constraints, ensuring compatibility with video animation:
		\begin{equation}
			\small
			A^i_j = G_A(p_{A,j}^i), \ p_{A,j}^{i+1} = R_A(p_{A,j}^i; A^i_j, V_j), \ A_j = A^n_j.
		\end{equation}
		
		Finally, audio tracks are synchronized with video clips to produce the final long-sequence video telling with synchronized visuals and sound: $ \hat{\mathcal{V}} = \{\{V_j, \ A_j\} \mid j = 1, 2, \dots, N\} $.

		\section{Experiment}
		
		To evaluate the effectiveness of \method, we benchmark it against three long video generation baselines: VGoT, Mora, and MovieAgent, and conduct detailed ablation studies. 
		We constructed a test set of 20 user prompts, each targeting a one-minute video for evaluation.
		We employ CLIP, Inception Score, and VBench~\cite{Vbench} for quantitative evaluation, and conduct a user study for subjective assessment. 
		Further experimental settings, metrics, and baseline details are provided in the Appendix.



		
		\subsection{Performance Comparisons}
		We evaluate each method via automatic metrics and user study.
		It's worth noting that \method~is the only framework that provides multimodal output -- videos with narratives and BGM.
		
		\subsubsection{Automatic Metrics}
		Table~\ref{Tab: automatic} presents the shot-level evaluation results for both keyframe generation and video animation. 
		\method \ achieves the highest scores in CLIP (\textit{34.22}) and Inception (\textit{12.81}) scores, attributed to the enhanced T2I quality enabled by the 3E Principle in the T2I process. 
		In video animation, \method \ slightly lags behind VGoT in subject consistency and background consistency but outperforms all baselines in temporal flickering, motion smoothness, and image quality. 
		This is partly because VGoT produces videos with limited dynamics and highly saturated visuals, which improves intra-shot consistency and aesthetic quality but at the cost of realism. 
		Overall, \method \ achieves the best comprehensive performance.

		\begin{table}[t]
			\scriptsize
			\centering
			\tabcolsep=0.367cm
			\begin{tabular}{l|ccc}
				\toprule
				Ablation & Narrative $\uparrow$ & Visual $\uparrow$ & User Align. $\uparrow$\\
				\hline
				w/o $E_{\mathrm{str}}$ & \cellcolor{lightgray}13.67 & 25.67 & 26.67 \\
				
				w/o $E_{\mathrm{con}}$ & 32.33 & \cellcolor{lightgray}18.67 & 26.67  \\
				
				w/o $E_{\mathrm{sty}}$ & \cellcolor{lightgray}20.00 & 26.67 & \cellcolor{lightgray}20.00   \\
				
				\hline
				\textbf{\method~(ours)} & {34.00} & {26.00} & {26.67}\\
				\bottomrule
			\end{tabular}
			\caption{
				\textbf{Ablation study on the Structure Reviewer $E_{\mathrm{str}}$, the Content Reviewer $E_{\mathrm{con}}$, and the Style Reviewer $E_{\mathrm{sty}}$}.
				$\uparrow$ indicates a higher value is more desired.
				\colorbox{lightgray}{Gray-shaded cells} indicate values that are significantly lower than the others.
			}
			\label{Tab: ablation-script}
		\end{table}
		
		\subsubsection{User Study}
		Table~\ref{Tab: user study} summarizes the voting results from 60 evaluators across multiple aspects of comparison among the baselines. 
		\method \ consistently outperforms all other methods in every evaluation dimension, with an average voting share of \textit{69.44\%}. 
		This highlights both the effectiveness of our architecture and the critical role of the 3E Principle. 

		\subsection{Ablation Study}
		We conduct an ablation study to demonstrate the impact of the 3E Principle across different modules of the framework.

		\subsubsection{Script Writing Guidelines}
		In the script writing stage, based on the Script Writing Guidelines, the Structure Reviewer focuses on how scenes and shots are segmented and arranged, the Content Reviewer ensures that each shot can fully convey its intended content, and the Style Reviewer oversees global pacing, logical flow, structural coherence, and alignment with user requirements. 
		Table~\ref{Tab: ablation-script} presents the effects of removing each reviewer from the pipeline. 
		As shown, excluding the Structure Reviewer significantly impairs narrative expressiveness, removing the Content Reviewer degrades visual quality, and omitting the Style Reviewer leads to a loss in both narrative alignment and consistency with user intent. Optimal script generation is achieved only when all three reviewers collaborate.
		
		\begin{table}[t]
			\scriptsize
			\centering
			\tabcolsep=0.08cm
				\begin{tabular}{l|l|ccc|cc}
					\toprule
					\multicolumn{2}{c|}{Ablation} & CLIP $\uparrow$ & Inception $\uparrow$ & Natural. $\uparrow$ & Sub. Cons. $\uparrow$ & Dyn. Deg. $\uparrow$\\
					\hline
					\multirow{2}{*}{T2I} & w/o 3E & 31.25 & 10.54 & \cellcolor{lightgray}72.00 & - & -\\
					& w 3E & {34.22} & 12.81 & 98.00 & - & - \\
					\hline
					\multirow{2}{*}{I2V} & w/o 3E  & 32.05 & 11.22 & \cellcolor{lightgray}60.00 & 3.0 & 44.58  \\
					& w 3E  & 34.53 & {13.22} & 92.50 & 95.72 & 35.62  \\
					
					\bottomrule
				\end{tabular}
				\caption{
					\textbf{Ablation study of the 3E Principle in the T2I and I2V Processes}. "Natural." refers to the Naturalness score, $\uparrow$ indicates a higher value is more desired. \colorbox{lightgray}{Gray-shaded cells} indicate values that are significantly lower.
				}
				\label{Tab: ablation-process}
			\end{table}
			
			\begin{table}[t]
				\scriptsize
				\centering
				\tabcolsep=0.16cm
				\begin{tabular}{l|c||l|c||l|c}
					\toprule
					Ablation & C. Rate $\uparrow$ & Ablation & C. Rate $\uparrow$ & Ablation & C. Rate $\uparrow$\\
					\hline
					w/o $E_{\textrm{shot}}$ & \cellcolor{lightgray}84.5 & w/o $E_{A}$ & \cellcolor{lightgray}92.0 & w/o $R_{A}$ & \cellcolor{lightgray}56.5 \\
					w $E_{\textrm{shot}}$ & 100.0 & w/o $E_{A}$ & 100.0 & w/o $R_{A}$ & 100.0 \\
					
					\bottomrule
				\end{tabular}
				\caption{
					\textbf{Ablation study on the Shot Reviewer $E_{\textrm{shot}}$, the Voice Reviewer $E_{A}$, and the Subtitle Refiner $R_{A}$}.
					"C. Rate" refers to "Compliance Rate". $\uparrow$ indicates a higher value is more desired. \colorbox{lightgray}{Gray-shaded cells} indicate values that are significantly lower.
				}
				\label{Tab: ablation-other}
			\end{table}

			\subsubsection{3E Principle in T2I and I2V Processes}
			Since the evaluators and refiners in both the T2I and I2V processes cannot operate independently, we conduct an ablation study to evaluate the overall impact of the 3E Principle on these modules. 
			The CLIP score, Inception score, Naturalness score (defined as the proportion of generated images or videos that appear physically plausible and naturally rendered), along with Subject Consistency and Dynamic Degree from the VBench metrics, are used to assess key aspects targeted by the T2I and I2V Evaluators.
			As shown in Table~\ref{Tab: ablation-process}, removing the 3E Principle leads to a noticeable drop in performance across both T2I and I2V outputs, particularly in terms of naturalness ($72.0\rightarrow98.0$ in T2I and $60.0\rightarrow92.5$ in I2V). 
			This highlights the critical role of the 3E Principle in improving generation quality.
			An exception is observed in the I2V process: the Dynamic Degree is higher without the 3E Principle ($44.58 > 35.63$). 
			This may be because high-dynamic videos are more prone to physical inconsistencies and instability, leading the agents to favor lower-dynamic, more stable results.

			\subsubsection{Shot Reviewer, Voice Reviewer, and Subtitle Refiner}
			As related, the Shot Reviewer assesses the completeness of the shot designs, the Voice Reviewer ensures the music availability and voice-emotion appropriateness of voice design, and the Subtitle Refiner shortens voice-over to match the duration limits of video clips. 
			We use Compliance Rate—defined as the proportion of outputs that fully adhere to their respective guidelines—to evaluate the contributions of these agents. 
			As shown in Table~\ref{Tab: ablation-other}, removing these agents leads to a noticeable drop in compliance, particularly in voice-over generation, which in turn negatively impacts downstream content generation. 
			These results underscore the necessity of the 3E Principle in shot designing, voice designing, and the speech audio generation.

			\section{Conclusion}
			In this work, we presented \method, a multi-agent end-to-end framework for long-sequence video storytelling. 
			By orchestrating multiple stages within a unified pipeline, 
			\method \ effectively addresses the key limitations of prior approaches — poor assistive capability, limited visual expressiveness, and incomplete outputs. 
			Central to our framework is the 3E Principle (Explore, Examine, Enhance), which enables iterative refinement across all stages, maximizing output quality. 
			Additionally, our proposed Script Writing Guidelines improve compatibility between narrative structure and generative tools. 
			Experimental results validate that \method \ significantly advances the assistive capability, visual quality, and expressiveness of long-sequence video storytelling.

			\section*{Limitations}

			In practical AI short film production, there are multiple ways to generate a shot—for example, using direct T2V, combining several I2V generations from different camera angles, or applying image editing tools and image generation models with stronger semantic control and compositional reasoning (e.g., nanobanana~\cite{nanobanana} and HunyuanImage 3.0~\cite{cao2025hunyuanimage}) to enforce identity and background consistency. However, this paper restricts long-sequence video storytelling to a T2I+I2V paradigm, and the scriptwriting guidelines are explicitly adapted to accommodate this pipeline. While this design simplifies the workflow, it also limits shot diversity, thereby constraining the expressive capacity of the generated storytelling.
			
			Moreover, current storylines lack interactions and dialogue between characters, and the absence of video editing mechanisms in the framework further degrades the cinematic quality of the final output.
			
			To address these limitations, future work will allow agents to autonomously select appropriate generation strategies and tools, and will incorporate character dialogues, scene-specific sound effects, and video editing functionalities, aiming to produce more expressive and coherent video storytelling.
			
			
			Another limitation of this work lies in the limited reliability of the T2I and I2V Evaluators. Current multimodal foundation models still exhibit limited accuracy in assessing visual content, resulting in suboptimal scoring performance when ranking candidate outputs. To mitigate this in practice, we incorporate additional evaluation operators—such as those from VBench for assessing prompt consistency and image quality—to assist the Evaluators. 
			However, to ensure fairness during experimental evaluation, these operators were not used. In future work, we plan to integrate a broader set of evaluation operators to enhance the overall assessment process.
			
			\section*{Ethical Considerations}
			
			While multi-agent frameworks like \method\ enable efficient and creative video storytelling, AI-generated media must be used responsibly. Because \method\ integrates multiple open-source models and commercial APIs, its outputs may inherit safety risks, biases, or limitations from these components. We encourage transparent disclosure of model provenance, dataset sources, and responsible use practices when employing \method\ in creative or research contexts.
			
			\paragraph{Potential Risks.}
			\method\ is designed as a research framework for controllable video generation and multimodal reasoning. However, its capacity to produce coherent, realistic long-sequence videos may be misused to create misleading or harmful synthetic content, such as disinformation, deepfakes, or fabricated narratives. In addition, because it relies on large pretrained models and APIs, \method\ may inadvertently propagate existing societal biases, cultural stereotypes, or skewed representations into generated stories. We emphasize that \method\ is intended solely for academic research and creative exploration under ethical guidelines, not for deployment in sensitive, deceptive, or adversarial applications.

			\bibliography{references}
			
			\clearpage

			\appendix
			
			\section{More Related Works}
			
			\subsection{Video Generation}
			Long video generation~\cite{LVDM,TATS-longv,wang2024leo} aims to produce longer videos than previous text-to-video (T2V)~\cite{modelscope,Lavie} and image-to-video (I2V)~\cite{EmuVideo,Dynamicrafter,I2VGen} models.
			Auto-regressive approaches~\cite{ART-V, Streamingt2v} extend videos by conditioning on earlier frames, yet often result in repetitive motions with limited narrative development. 
			Attempts to directly integrate storytelling and video generation~\cite{Nuwa-xl, VideoGen-of-Thought} suffer from poor visual quality and unstructured video composition, leading to subpar viewer experiences.
			Meanwhile, commercial models have achieved notable success in both T2V~\cite{Mochi, Hunyuanvideo, Sora} and I2V generation~\cite{Veo2, Luma, hailuo, wanx}, and some ~\cite{Kling, Gen3, Gen4} support temporal extension.

			\subsection{Image Generation}
			Recent advances in image generation~\cite{Storygan,Attngan,Stackgan, DALLE} have been propelled by deep learning techniques.
			To fulfill the task of visual storytelling, some works~\cite{StoryGen, SEED-story, Make-a-story} generate image sequences from text and image-caption pairs. 
			To ensure subject consistency~\cite{TextualInversion}, DreamBooth~\cite{Dreambooth} adopts subject-specific fine-tuning, while GPT-Image-1~\cite{gptImage} introduces context-aware generation for identity preservation. 
			Recent models such as NanoBanana~\cite{nanobanana} and HunyuanImage 3.0~\cite{cao2025hunyuanimage} further advance high-fidelity and instruction-following generation, offering stronger semantic control and compositional reasoning capabilities.
			In this work, \method \ leverages off-the-shelf models to generate keyframes for each shot, which serve as the basis for subsequent video animation.
			
			\subsection{AI Agents}
			AI agents, empowered by large language models~\cite{GPT,Llama,Llama2}, have shown broad applicability in reasoning, planning, web interaction, and visual generation. 
			General-purpose LLMs underpin agent frameworks~\cite{autogpt, Lumos, Camel, Metagpt, GenerativeAgents}, while multimodal models~\cite{GPT4o,Gemini} extend agent capabilities across modalities~\cite{MM-REACT, auto-ui, SeeAct, Jarvis-1, Llava-plus, Genartist, CCA, MotionAgent}. 
			In video generation, systems~\cite{WorldGPT, Mora, AesopAgent} explore world modeling, multi-agent coordination, and retrieval-augmented synthesis, yet often yield static content with limited transitions. 
			Hierarchical pipelines~\cite{Dreamfactory, MovieAgent} introduce script-to-shot workflows but still rely on manual scripting and LoRA tuning.

			\section{Explanation and Discuss about the Script Writing Guideline}
			Due to space limitations in the main text, we provided only a brief overview of the Script Writing Guideline. In this section, we offer a more detailed explanation of the guideline and elaborate on the rationale behind each design constraint.
			
			In the Script Writing Guideline, "\textit{Scriptwriters should minimize transitions between tightly connected spatial locations}" refers to avoiding consecutive shots set in spatially adjacent backgrounds—for example, the first shot showing the character entering a room by going through a door, and the next shot showing the character from the front inside the room. Such transitions can break immersion due to background inconsistencies, such as mismatched door appearances between shots.
			
			The script writing guidelines are based on practical experience to improve compatibility with generation tools. E.g., "\textit{not to place the same character in consecutive shots}" stems from the fact that such sequences often involve multi-angle cuts in the same setting. Given current model limitations in maintaining background consistency, this approach is not yet reliably achievable.
			
			The constraints in the script writing guideline are based on the following empirical observations:
			
			\noindent $\bullet$ \textbf{\textit{Avoid placing successive shots in the same background}}: We measured VBench background consistency for adjacent shots sharing the same background and found it to be only 78.42, compared to 95.88 of single-shot. This justifies avoiding such situations.
			
			\noindent $\bullet$ \textbf{\textit{Minimize transitions between tightly connected spatial locations}}: For pairs of shots with strong spatial connections, we used GPT to assess whether their backgrounds were spatially related as described in the script. Only 26 out of 100 pairs were judged to be consistent, indicating such transitions should be avoided.
			
			\noindent $\bullet$ \textbf{\textit{Avoid complex actions within a single shot}}: Shots with complex actions achieved a VBench overall consistency of 23.25, compared to 27.76 for simple actions, suggesting simpler actions improve generation reliability.
			
			\noindent $\bullet$ \textbf{\textit{Avoid fine-grained visual elements within a single shot}}: Shots containing fine-grained visual elements scored 22.68 in VBench overall consistency, while those without such elements scored 27.76. Thus, they should be avoided.
			
			\noindent $\bullet$ \textbf{\textit{Align style with user's quest}}: Using GPT to judge alignment with user instructions, only 82 out of 100 scripts were fully aligned without constraints, compared to 100 with constraints in place.
			
			\noindent $\bullet$ \textbf{\textit{Ensure script structural completeness}}: A complete structure is essential for enabling subsequent stages of the generation pipeline.

			\begin{table*}[ht]
				\scriptsize
				\centering
				\tabcolsep=0.24cm
				\begin{tabular}{l|llll}
					\toprule
					\textbf{Stage} & \textbf{Agent/Model} & \textbf{Role} & \textbf{Reviewer}  & \textbf{Refiner}   \\
					\midrule
					Script Writing & Scriptwriter & Write script & Structure Reviewer, \\
					& & & Content Reviewer, & - \\
					& & & Style Reviewer  \\ 
					\hline
					Shot Designing & Shot Designer & Design each shot of the script & Shot Reviewer & - \\
					\hline
					Keyframe Generation & T2I Prompt Generator & Generate T2I prompt for each shot & - & - \\
					\hline
					Keyframe Generation,  & T2I Models & Text-to-image generation for each shot/character & T2I Evaluator & T2I Refiner \\
					Character Modeling \\
					\hline
					Video Animation,  & I2V Prompt Generator & Generate I2V prompt for each shot/character & - & - \\
					Character Modeling \\
					\hline
					Video Animation & I2V Models & Image-to-video generation for each shot & I2V Evaluator & I2V Refiner \\
					\hline
					Character Modeling & Character Prompt Generator & Generate T2I and I2V prompts for each character & - & - \\
					\hline
					Character Modeling & Caption Generator & Generate captions for each character images & - & - \\
					\hline
					Audio Generation & Voice Designer & Choose voice ID, emotion, and BGM & Voice Reviewer & - \\
					\hline
					Audio Generation & T2A Model & Text-to-audio generation for each shot & - & Subtitle Refiner \\
					\bottomrule
				\end{tabular}
				\caption{
					\textbf{Summary of Agents.} 
				}
				\label{tab: agent summary}
			\end{table*}

			\begin{table}[t]
				\scriptsize
				\centering
				\tabcolsep=0.4cm
				\begin{tabular}{l|ccc}
					\toprule
					\textbf{Methods} & FID$\downarrow$ & Inception$\uparrow$ & CLIP$\uparrow$\\
					\midrule
					DreamFactory & $7.03^{\dag}$ & $169.71^{\dag}$ & $30.92^{\dag}$\\ 
					\textbf{\method} & - & 12.81 & 34.22  \\
					\bottomrule
				\end{tabular}
				\caption{
					Compare with DreamFactory~\cite{Dreamfactory}. $\dag$ denotes the reported score in the paper of the baseline method.
				}
				\label{tab: Dreamfactory}
			\end{table}
			
			\begin{table}[t]
				\scriptsize
				\centering
				\tabcolsep=0.25cm
				\begin{tabular}{l|cccc}
					\toprule
					\textbf{Methods} & CLIP$\uparrow$ & Inception$\uparrow$ & AS$\uparrow$ & CLIP-sim$\uparrow$\\
					\midrule
					MovieDreamer &  $19.52^\dag$ & $8.64^\dag$ & $6.05^\dag$ & $0.70^\dag$\\ 
					\textbf{\method} & 34.22 & 12.81 & 7.25 & 0.71  \\
					\bottomrule
				\end{tabular}
				\caption{
					Compare with MovieDreamer~\cite{MovieDreamer}. $\dag$ denotes the reported score in the paper of the baseline method. "AS" denotes Aesthetic Score used in~\cite{schuhmann2022laion}.
				}
				\label{tab: MovieDreamer}
			\end{table}

			\begin{table}[t]
				\scriptsize
				\centering
				\tabcolsep=0.06cm
				\begin{tabular}{l|cccc}
					\toprule
					\textbf{Methods} & Aesthetic$\uparrow$ & Image Quality$\uparrow$ & Consistency (avg.)$\uparrow$ & P. Consistency$\uparrow$\\
					\midrule
					LCT & $60.79^\dag$ & $67.44^\dag$ & $95.65^\dag$ & $30.14^\dag$\\ 
					\textbf{\method} & 63.17 & 72.91 & 95.92 & 34.22  \\
					\bottomrule
				\end{tabular}
				\caption{
					Compare with LCT~\cite{LCT}. $\dag$ denotes the reported score in the paper of the baseline method.
					“Consistency (avg.)” represents the average score of subject and background consistency.
					"P. Consistency" denotes Prompt Consistency.
				}
				\label{tab: LCT}
			\end{table}
			
			\section{Implementation Details}
			All agents in our framework are implemented using AutoGen~\cite{Autogen} and powered by GPT-4o~\cite{GPT4o}.
			The T2I model pool comprises FLUX.1~\cite{FLUX}, Stable Diffusion 3.5~\cite{stable_diffusion}, and GPT-Image~\cite{gptImage}, while the I2V model pool includes Veo2 \cite{Veo2}, Gen-3~\cite{Gen3}, Gen-4~\cite{Gen4}, LUMA~\cite{Luma}, HunyuanVideo-I2V~\cite{Hunyuanvideo}, and Wan2.1~\cite{wanx}. 
			We use MUDI~\cite{MUDI} for LoRA training and Hailuo \cite{hailuo} for the T2A generation.
			The maximum number of 3E iterations is set to 4 for script writing, shot designing, and voice designing; 2 for the T2I process; 1 for the I2V process; and 5 for subtitle refining.
			As the first work targeting end-to-end long-sequence video storytelling, no suitable benchmark dataset exists. 
			Therefore, we constructed a test set of 20 user prompts, each targeting a one-minute video for evaluation.

			\begin{table*}[t]
				\scriptsize
				\centering
				\begin{tabular}{l|l}
					\toprule
					\textbf{Metric} & Question\\
					\midrule
					Narrative Expressiveness & Which video performed best overall in terms of narrative engagement, story coherence, and viewing experience?\\ 
					Visual Quality & Which video was the most visually expressive in terms of emotion, atmosphere, and cinematic feel?  \\
					User Prompt Alignment & Which video best aligns with the original user prompt (e.g., character setup, plot elements, or keywords)? \\
					Character Consistency & Which video had the most consistent character appearances across scenes? (clothing, hairstyle, facial features) \\
					Character Naturalness & Which video had the most natural and correct character generation? Consider anatomical realism, physical motion plausibility, \\
					& and the absence of duplicate or mistakenly generated characters. \\
					Background Consistency & Which video had the most consistent and logically coherent background style across shots? \\
					Background Realism & Which video had the most natural background elements and physically plausible movements?\\
					\bottomrule
				\end{tabular}
				\caption{
					Metrics and Questions for User Study.
				}
				\label{tab: user question}
			\end{table*}
			
			\begin{table*}[t]
				\scriptsize
				\centering
				\tabcolsep=0.2cm
				\begin{tabular}{l|cc|cccccc}
					\toprule
					\multirow{2}{*}{Method} & \multicolumn{2}{c}{Keyframe Generation} & \multicolumn{6}{|c}{Video Animation} \\
					\cline{2-9}
					& CLIP $\uparrow$ & Inception $\uparrow$ & T. Flick. $\uparrow$ & M. Smooth. $\uparrow$ & Sub. Cons. $\uparrow$ & Bg. Cons. $\uparrow$ & Aesthetic $\uparrow$ & I. Quality $\uparrow$\\
					\hline
					VGoT~\cite{VideoGen-of-Thought} & 22.32 & 8.41 & 99.00 & 99.28 & \textbf{98.98} & \textbf{98.76} & \textbf{79.95} & 64.62\\
					
					Mora~\cite{Mora} & 34.02 & 12.70 & 97.25 & 99.21 & 94.58 & 95.01 & 64.41 & 71.39\\
					
					MovieAgent~\cite{MovieAgent} & 31.64 & 10.69 & 97.05 & 99.18 & 93.44 & 94.62 & 63.38 & 91.25  \\
					
					\hline
					\textbf{\method~(ours)} & \textbf{34.15} & \textbf{12.79} & \textbf{99.08} & \textbf{99.52} & 95.54 & 96.10 & 64.05 & \textbf{72.68}\\
					\bottomrule
				\end{tabular}
				\caption{
					Compare with baselines on MovieBench and ViStoryBench. 
				}
				\label{tab: moviebench}
			\end{table*}

			To standardize intermediate outputs for downstream processes, we introduced a Script Consolidator agent in the Script Writing stage. This agent formats the script after each 3E iteration to ensure consistency. 
			In both the T2I and I2V processes, due to limitations on the number of images that can be processed simultaneously, the T2I Evaluator and I2V Evaluator adopt a batched evaluation strategy. 
			Following this, T2I Selector and I2V Selector agents are employed to select the best candidates based on all evaluation results. Specifically, the I2V Evaluator samples 8 evenly sampled frames from each video candidate for evaluation.
			
			The T2I process in the Shot Designing stage mirrors that of Character Modeling, with the difference being that the T2I model pool in the Shot Designing stage incorporates the LoRA models trained during Character Modeling. 
			We use MUDI~\cite{MUDI} for LoRA training, which supports multi-subject generation, allowing a single LoRA to serve an entire script.
			
			Given that different models in the I2V model pool require reference images at different resolutions, keyframes are center-cropped and resized before I2V execution according to model-specific requirements. 
			To ensure consistency, all generated videos are center-cropped and resized to 1280$\times$720 resolution.
			
			The T2I model pool includes the open-source models FLUX~\cite{FLUX} and Stable Diffusion 3.5~\cite{SD3.5}, as well as the API-based GPT-Image~\cite{gptImage}. 
			The I2V model pool comprises open-source models Hunyuan~\cite{Hunyuanvideo} and Wan2.1~\cite{wanx}, along with API-based models Gen3~\cite{Gen3}, Gen4~\cite{Gen4}, LUMA~\cite{Luma}, and Veo2~\cite{Veo2}. 
			For T2A, we use the Minimax Hailuo API~\cite{hailuo}.
			All experiments were conducted on two NVIDIA H100 GPU.
			
			More samples generated by \method  ~framework are shown in Figure.~\ref{fig: tail1} and Figure.~\ref{fig: tail2}.

			\subsection{\textbf{Metrics}}
			we constructed a test set of 20 user prompts, each targeting a one-minute video for evaluation.
			We employ CLIP and Inception Score to evaluate prompt consistency and generation quality in the keyframe generation stage, and use VBench~\cite{Vbench} to assess video animation performance.
			Additionally, a user study is conducted to assess real user preferences and validate the practical effectiveness of MAViS and baselines.
			For all evaluation metrics, higher values indicate better performance.
			
			\paragraph{Compliance Rate}
			Compliance Rate is an objective and transparent metric defined as the fraction of outputs that are fully compliant with the guidelines (Line 596).
			A “fully compliant” output is defined as follows:
			$\bullet$ Shot Designing: All seven fields are present $+$ in the correct format.
			$\bullet$ Voice Designing: Downloadable BGM $+$ consistent voice ID $+$ correct format.
			$\bullet$ Subtitle Refining: Narration duration $\leq$ video duration.
			
			If any of the above rules are violated, the workflow will produce an error. Therefore, the Compliance Rate directly measures the necessity and effectiveness of the 3E principle across these three stages, and there is no better metric that can faithfully reflect this property.

			\subsection{\textbf{Baselines}}
			We compare \method \ with three long video generation baselines: VGoT, Mora, and MovieAgent. 
			Among them, Mora and MovieAgent adopt a modular design and are equipped with the same image and video generation models as \method, and are also powered by GPT-4o. 
			Since Mora and MovieAgent do not support script generation, we provide them with scripts produced by our script writing stage. 
			For MovieAgent, which requires manual training of LoRA models, we generate character images based on the provided scripts and train LoRA models for each script.

			\section{Compare with More Baselines}
			In our experiments, we conducted direct comparisons with three open-source baselines: VGoT~\cite{VideoGen-of-Thought}, Mora~\cite{Mora}, and MovieAgent~\cite{MovieAgent}. Subsequently, we compared our results with the reported performances of non-open-source baselines.
			
			We first compare our method with DreamFactory~\cite{Dreamfactory}. Due to the lack of real videos in automatic long-sequence video storytelling, it is not feasible to compute the FID metric. As shown in Table~\ref{tab: Dreamfactory}, \method~achieves higher scores across most metrics; however, there is a substantial gap in Inception Score. We adopted a widely used implementation of the Inception Score, which may differ from the version used by DreamFactory, potentially leading to differences in score range. Since the baseline is not open-sourced, we are unable to verify the exact cause of this discrepancy.

			Next, we compare our method with MovieDreamer~\cite{MovieDreamer}. As shown in Table~\ref{tab: MovieDreamer}, \method~outperforms this baseline across all evaluation metrics.

			Subsequently, we compare our method with LCT~\cite{LCT}, using evaluation metrics sourced entirely from VBench. As shown in Table~\ref{tab: LCT}, \method~consistently outperforms this baseline across all metrics.

			\section{User Study}
\begin{figure*}[t]
				\centering
				\includegraphics[width=2\columnwidth]{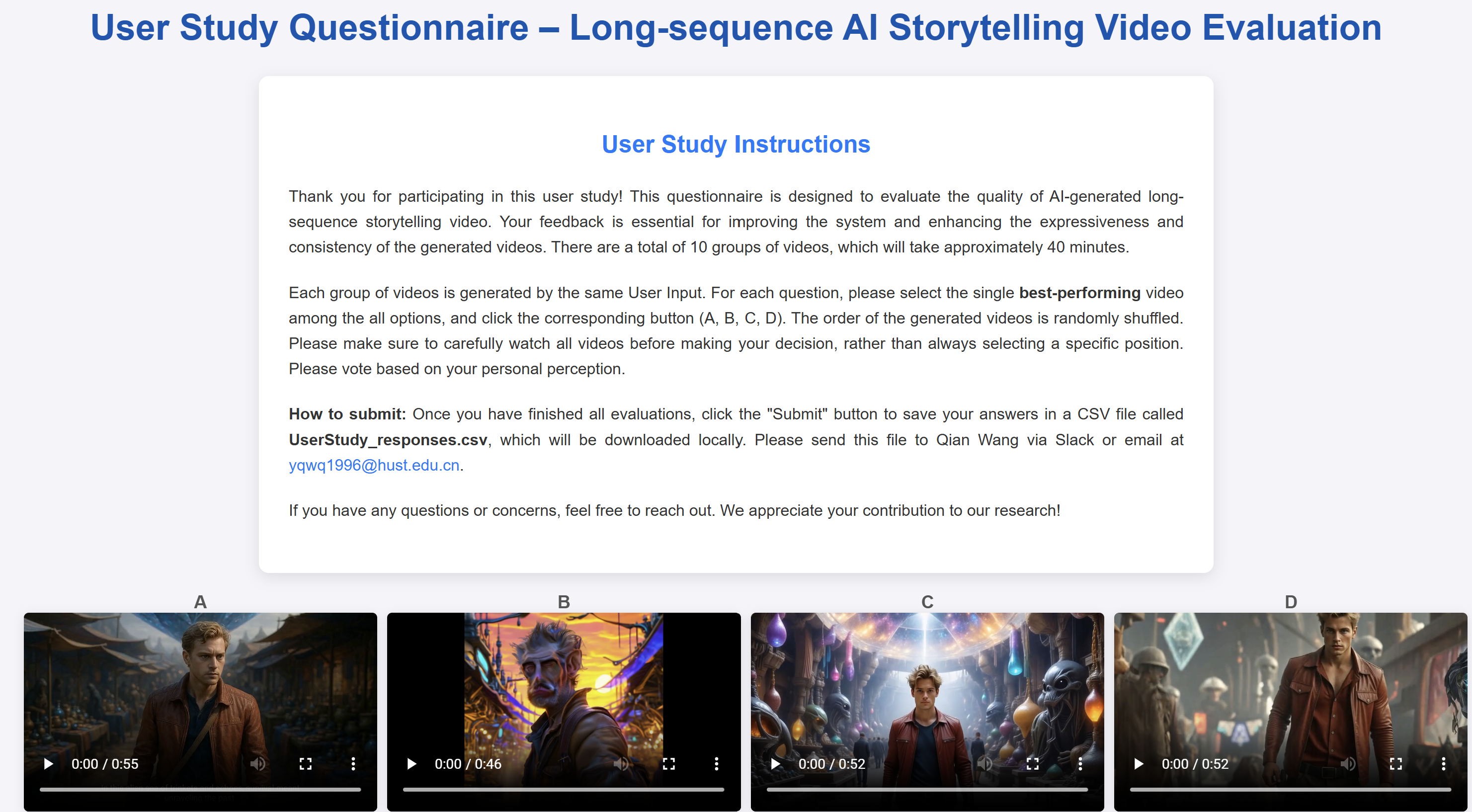}
				\caption{
					Screenshot of the user study HTML interface with instructions and generated video candidates.
				}
				\label{fig: user study html1}
			\end{figure*}

\begin{figure*}[ht]
				\centering
				\includegraphics[width=2\columnwidth]{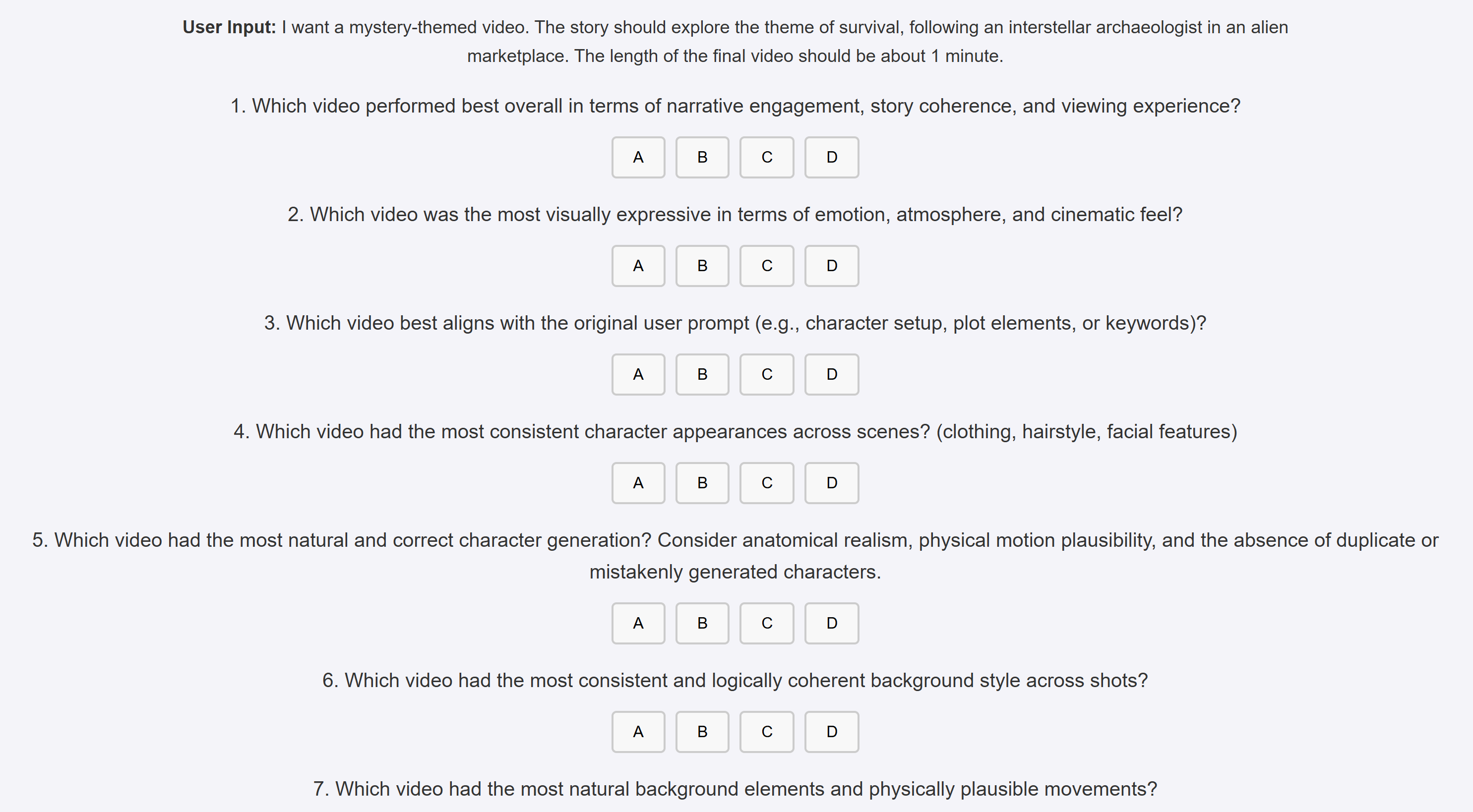}
				\caption{
					Screenshot of the user study HTML interface with seven evaluation questions.
				}
				\label{fig: user study html2}
			\end{figure*}

\begin{figure*}[ht]
				\centering
				\includegraphics[width=2\columnwidth]{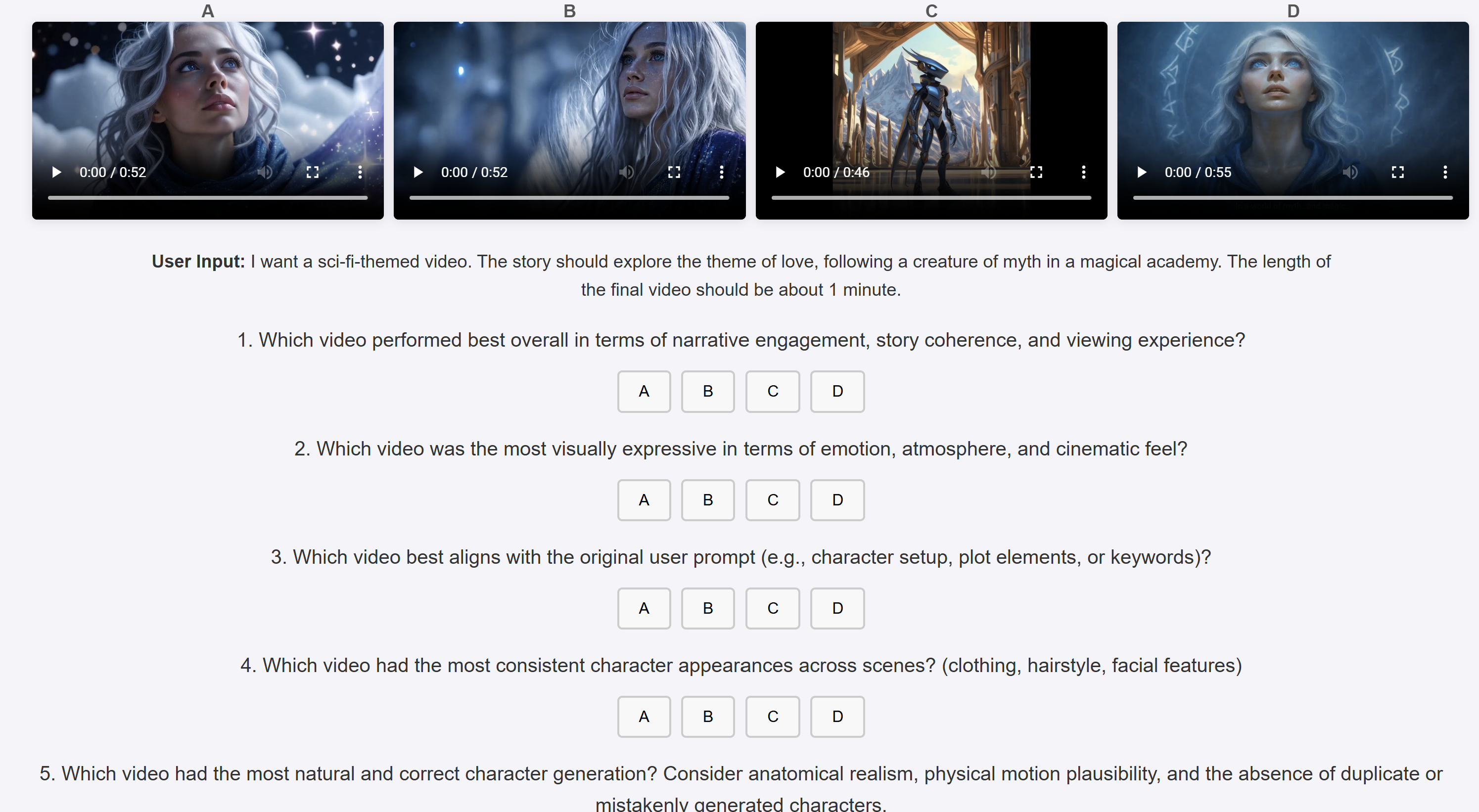}
				\caption{
					Screenshot of the user study HTML interface with generated video candidates and seven evaluation questions.
				}
				\label{fig: user study html3}
			\end{figure*}

			In the User Study, participants were asked to vote for the best video in each group based on seven evaluation criteria. 
			As listed in Table~\ref{tab: user question}, the criteria include: Narrative Expressiveness, Visual Quality, User Prompt Alignment, Character Consistency, Character Naturalness, Background Consistency, and Background Realism. 
			Prior to voting, detailed guidelines for each metric were provided to ensure consistency and reliability in evaluation. 
			Some screenshots of the user study HTML interface is shown in Figures~\ref{fig: user study html1}, ~\ref{fig: user study html2}, and~\ref{fig: user study html3}.

			\section{Evaluation on the Existing Dataset}
			
			We are establishing an end-to-end agentic system from users' single line of story theme to final pixels plus other multimodalities. However, existing long video benchmarks, MovieBench~\cite{moviebench} and ViStoryBench~\cite{vistorybench}, focus only on benchmarking T2I and I2V models', not on the entire agentic system. In particular, they ignore the script generation and scripts' compatibility with existing T2I and I2V models' capabilities.
			
			Nevertheless, we sampled three examples (each consisting of 10 shots) from both MovieBench and ViStoryBench. As shown in Table~\ref{tab: moviebench}, MAViS consistently outperforms the compared methods.

			\section{Quality-Efficiency Analysis}
			
			In the stages where the 3E principle is applied, we evaluated the impact of the maximum iteration number ($I$) on generation quality and time using the same metrics as in our paper.It's worth noting that C. Rate refers to Compliance Rate, defined as the proportion of outputs that fully adhere to their respective guidelines. It is used to measure the success rate of each stage's generation.
			
			\begin{table}[t]
				\scriptsize
				\centering
				\tabcolsep=0.2cm
				\begin{tabular}{l|ccccc}
					\toprule
					$I$ & Narrative & Visual & User Align. & C.Rate & Time(Second)\\
					\midrule
					0 & 12.58 & 18.44 & 19.56 & 74.00 & 9.80 \\ 
					1 & 25.32 & 22.85 & 24.40 & 86.00 & 20.90 \\
					2 & 32.54 & 24.82 & 26.49 & 94.50 & 28.24 \\
					3 & 33.48 & 25.60 & 26.52 & 98.50 & 36.13 \\
					4 & 34.00 & 26.00 & 26.67 & 100.00 & 41.58 \\
					\bottomrule
				\end{tabular}
				\caption{
					Quality-Efficiency Analysis on scriptwriting.
				}
				\label{tab: Q-E scriptwriting}
			\end{table}

			\begin{table}[t]
				\scriptsize
				\centering
				\tabcolsep=0.2cm
				\begin{tabular}{l|ccccc}
					\toprule
					$I$ & CLIP & Inception & Natural. & C.Rate & Time(Second)\\
					\midrule
					0 & 31.25 & 10.54 & 72.00 & 68.00 & 175.19 \\
					1 & 32.25 & 11.30 & 84.00 &81.00 & 381.32 \\
					2 & 34.22 & 12.81 & 98.00 & 92.50 & 557.47 \\
					3 & 34.26 & 12.78 & 98.50 & 96.00 & 602.06 \\
					\bottomrule
				\end{tabular}
				\caption{
					Quality-Efficiency Analysis on Keyframe Generation.
				}
				\label{tab: Q-E T2I}
			\end{table}

			For script writing, as shown in Table~\ref{tab: Q-E scriptwriting}, the performance is already satisfactory at $I=2$. Increasing $I$ further can improve generation quality with minimal impact on overall efficiency.

			\begin{table}[t]
				\scriptsize
				\centering
				\tabcolsep=0.06cm
				\begin{tabular}{l|ccccccc}
					\toprule
					$I$ & CLIP & Inception & Natural. & Sub. Cons. & Dyn. Deg. & C.Rate & Time(Minute)\\
					\midrule
					0 & 32.05 & 11.22 & 60.00 & 93.00 & 44.58 & 72.50 & 61.17 \\
					1 & 34.53 & 13.22 & 92.50 & 95.72 & 35.62 & 86.00 & 124.21 \\
					2 & 34.62 & 13.31 & 95.00 & 95.51 & 35.68 & 94.00 & 176.77 \\
					3 & 34.35 & 13.18 & 96.50 & 95.46 & 35.54 & 98.00 & 215.38 \\
					\bottomrule
				\end{tabular}
				\caption{
					Quality-Efficiency Analysis on Video Animation.
				}
				\label{tab: Q-E I2V}
			\end{table}

			\begin{table}[ht]
				\scriptsize
				\centering
				\tabcolsep=0.08cm
				\begin{tabular}{l|cc|cc|cc}
					\toprule
					& \multicolumn{2}{c|}{Shot Designing} & \multicolumn{2}{c|}{Voice Designing} & \multicolumn{2}{c}{Subtitle Refining} \\
					$I$ & C.Rate & Time(Second) & C.Rate & Time(Second) & C.Rate & Time(Second)\\
					\midrule
					0 & 84.50 & 24.34 & 92.00 & 5.62 & 56.50 & 1.58 \\
					1 & 89.00 & 73.49 & 95.00 & 13.48 & 86.00 & 2.83 \\
					2 & 94.00 & 122.352 & 98.50 & 26.19 & 92.50 & 3.95 \\
					3 & 97.00 & 161.21 & 97.00 & 161.21 & 98.00 & 4.36 \\
					4 & 98.50 & 195.86 & 99.00 & 29.84 & 99.50 & 4.84 \\
					5 & 100.00 & 231.60 & 100.00 & 31.68 & 100.00 & 4.89 \\
					\bottomrule
				\end{tabular}
				\caption{
					Quality-Efficiency Analysis on shot designing, voice designing, and subtitle refining.
				}
				\label{tab: Q-E others}
			\end{table}
			
			For keyframe generation, as shown in Table~\ref{tab: Q-E T2I}, the performance is already satisfactory at $I=2$. Increasing $I$ further can enhance generation quality, with only a slight impact on overall efficiency.

			For video animation, as shown in Table~\ref{tab: Q-E I2V}, the performance is already satisfactory at $I=2$. While increasing $I$ can further improve quality, it has a significant impact on overall efficiency. Therefore, we set $I=1$ for video animation in our experiments.

			For shot designing, voice designing, and subtitle refining, as shown in Table~\ref{tab: Q-E others}, increasing $I$ improves generation quality with minimal impact on overall efficiency. However, since errors in these stages can critically affect the entire generation process, we set $I=5$ for these tasks in our experiments.
			
			Additionally, since iterations may terminate before reaching the maximum iteration number, the generation time does not scale linearly with $I$.

			\begin{table}[t]
				\scriptsize
				\centering
				\tabcolsep=0.2cm
				\begin{tabular}{l|ccccccc}
					\toprule
					Model/API & Cost & Time overhead & Duration & GPU \\
					\midrule
					Veo 2 & 2.8\$ & 31.7s & 8s & - \\
					Gen 3 & 0.5\$ & 23.7s & 10s & - \\
					Gen 4 & 0.5\$ & 40.8s & 10s & - \\
					Luma & 1.27\$ & 36.2s & 5s & - \\
					Hunyuan-I2V & - & 28.7min & 5s & 1 H100 \\
					Wan 2.1 & - & 36.3min & 3s & 1 H100 \\
					\bottomrule
				\end{tabular}
				\caption{
					Real-world Cost per Shot.
				}
				\label{tab: cost}
			\end{table}

			\section{Generation Time Overhead.}
			
			Using all models in the model pool, the average time for MAViS to generate a long video on two H100 GPUs is 13.63 hours. In contrast, it will take our internal artists 5 days to complete a pipeline similar to ours: data generation, evaluation, selection, and composition. The process can be significantly accelerated if only API-based I2V models are used.
			The resources required for a single video model to generate one 720p shot are as shown in Table~\ref{tab: cost}.

			\section{Most Common Issues and Success Rate}
			We report the most common issues and success rates across all the stages that apply the 3E principle during generation:
			
			$\bullet$ Script writing: Adjacent shots are set in the same or strongly connected locations, violating the script writing guidelines.
			
			$\bullet$ Shot designing: Incomplete content, such as insufficient background detail.
			
			$\bullet$ T2I: Facial and limb distortions.
			
			$\bullet$ I2V: Abnormal character movements.
			
			$\bullet$ Voice designing: BGM download failures.
			
			$\bullet$ Subtitle refinement: The voice duration exceeds the clip length.
			
			Using criteria such as avoiding visual tearing, distortion, physically implausible motion, ID inconsistency, and style inconsistency as overall success measures, MAViS achieves 0.8621, compared to 0.4875 without the 3E principle.

			\begin{figure*}[t] 
				\centering
				\scriptsize
				\begin{tcolorbox}[
					title={Scriptwriter System Message},
					fonttitle=\ttfamily\bfseries,
					colframe=black!70!white,
					colback=black!5!white,
					fontupper=\small\ttfamily,
					width=\textwidth,
					before upper={\setlength{\parskip}{8pt}\setlength{\parindent}{0pt}}
					]
					
					{\ttfamily
						
						Role Definition: \\
						You are a professional **"AI Microfilm Scriptwriting Expert"**, specializing in writing **technically compliant shot-by-shot scripts** for AI video generation tools. The scripts must be **visually concise**, **logically coherent**, **engaging**, and **aligned with user-specified themes**, while ensuring **completely adherence to technical constraints** of AI video generation systems.
						
						---
						
						Task Objective: \\ 
						Generate shot-by-shot scripts suitable for AI micro-video generation tools that meet the following criteria:\\
						\hspace*{1em} - **Emotional**: Evokes emotional resonance such as family bonds, memories, or a sense of belonging.\\
						\hspace*{1em} - **Concise**: Slow-paced and restrained storytelling with clean, uncluttered visuals.\\
						\hspace*{1em} - **Logical**: Natural transitions between shots and scenes, with clear narrative clues.\\
						\hspace*{1em} - **Technically feasible**: All content must be executable within AI generation constraints, avoiding technical conflicts.
						
						---
						
						Workflow:\\
						1. **Clarify the Theme**  \\
						\hspace*{1em} - The script theme must align with the user's request. If unspecified, prioritize touching, emotional, and resonant themes, such as family, homecoming, or nostalgia.\\  
						\hspace*{1em} - Unless explicitly requested otherwise, **default to Western cultural and character backgrounds**; avoid Asian settings or characters.\\
						2. **Script Structure **\\  
						\hspace*{1em} - Each script must include:\\
						\hspace*{1em} - **Title** \\
						\hspace*{1em} - **Character Definitions** (see section below for format) \\
						\hspace*{1em} - **Shot-by-shot Script**: Minimum of 10 shots. If there's a final shot, name it `"Ending"` instead of "Shot 14" etc.\\
						\hspace*{1em} - Each shot must include:\\
						\hspace*{2em} - **Character**: Name of the character in frame, or "None" if no identifiable character appears.\\
						\hspace*{3em} - Always use the exact name defined in the character setup.\\
						\hspace*{3em} - Do not abbreviate, shorten, or alter the character name.\\
						\hspace*{2em} - **Shot Content**: Character action + background + framing info. \\
						\hspace*{2em} - **Subtitle/Voice-over**: Must be deliverable within 5 seconds, and match the tone and style of the story.\\
						\hspace*{3em} - Important: Do not use dashes ("—") in subtitle. Replace all such instances with ellipses ("...").\\
						
						---
						
						Character Definition Rules:\\
						\hspace*{1em} - Character names must follow these rules:\\
						\hspace*{2em} - Use only a simple first name (e.g., "Ethan", "Astra").\\
						\hspace*{2em} - Do not include last names, middle names, or titles/descriptors (Incorrect: "Ethan Carter", "The Guide". Correct: "Ethan", "Astra").\\
						\hspace*{1em} - Each character must be defined with 1-2 sentences, including:  \\
						\hspace*{2em} - Approximate age (e.g., "around 30 years old")  \\
						\hspace*{2em} - Ethnicity or race (e.g., "Caucasian")  \\
						\hspace*{2em} - Appearance (e.g., hairstyle, skin tone, height) \\ 
						\hspace*{2em} - Outfit style (e.g., "gray puffer jacket + jeans")  \\
						\hspace*{2em} - Notable features (e.g., "stoic gaze", "carries a backpack")\\
						Background characters don't require individual profiles—just mention their presence in scene descriptions.
						
						---

					}
					
				\end{tcolorbox}
				\caption{\textbf{Scriptwriter System Message.}}
				\label{fig:sysm_scriptwriter1}
			\end{figure*}

			\begin{figure*}[h] 
				\centering
				\scriptsize
				\begin{tcolorbox}[
					title={Scriptwriter System Message},
					fonttitle=\ttfamily\bfseries,
					colframe=black!70!white,
					colback=black!5!white,
					fontupper=\small\ttfamily,
					width=\textwidth,
					before upper={\setlength{\parskip}{8pt}\setlength{\parindent}{0pt}}
					]
					
					{\ttfamily
						
						Technical Constraints for AI Video Generation:
						
						Shot Content \& Action Rules:\\
						\hspace*{1em} - Each shot corresponds to a ~5-second video.\\
						\hspace*{1em} - Each shot must include **only one primary action**. Avoid chaining actions; **complex actions should be broken into multiple shots**.\\
						\hspace*{2em} - Incorrect: "He opens the door, sees his mother, eyes turning red" → Too many linked actions  \\
						\hspace*{2em} - Correct: Separate into "opens door," "mother's back in close-up," "his eyes turning red"\\
						\hspace*{1em} - Interaction between characters should be simple (e.g., hug, holding hands), avoiding complex physical interactions.\\
						\hspace*{1em} - No smoking or drinking behaviors.
						
						Scene Control \& Repetition Avoidance:\\
						\hspace*{1em} - Avoid using visually indistinct or similar settings (e.g., multiple shots in undifferentiated desert terrain without time/weather/landform distinction). Shots that differ in time of day, terrain type, or atmosphere are acceptable.\\
						\hspace*{1em} - Avoid strong spatial interactions between consecutive shots (e.g., entering/exiting doors, moving between rooms). Such transitions can break immersion due to background inconsistencies, such as mismatched door appearances between shots.
						
						Props and Visual Limitations:\\
						\hspace*{1em} - Do not show specific textual content (e.g., on photos, screens, paper) unless it's a **close-up** of the object.\\
						\hspace*{1em} - Avoid complex visual details (e.g., phone screens, photos, mirrors) unless it's a close-up.\\
						\hspace*{1em} - For still-life close-ups, **specify the background** (e.g., "close-up of a soup bowl on a wooden table"), and avoid blurred backgrounds.    
						
						Characters and Presentation:\\
						\hspace*{1em} - Do not show the **same character in two adjacent shots** (e.g., Shot 1 and Shot 2), unless a different character or a non-character shot (e.g., object-only, background-only, or non-identifiable body part) appears between them.\\
						\hspace*{2em} - **Voice-over mentioning a character does not count as character appearance.** Only shots where the character in the **Character** part count toward this rule.\\
						\hspace*{2em} - Switching between different characters (e.g., Shot 1 with character Ethan → Shot 2 with character Astra) **is allowed** and does not violate this rule.\\
						\hspace*{2em} - The same character in none-adjacent shots (e.g., Ethan in Shot 1, Shot 3, and Shot 5) **is allowed** and does not violate this rule.\\
						\hspace*{2em} - If a shot contains only a non-identifiable body part (e.g., hand close-up), it should be marked as "None" and does not violate the adjacent character appearance rule.\\
						\hspace*{1em} - Each shot must clearly specify the **character** being filmed. If the shot does not include a recognizable character or only includes a close-up of a body part that cannot identify the character (e.g., hand, foot, leg—not the head), use `"None"`. \\
						\hspace*{1em} - Character clothing must remain consistent throughout the script—**do not change outfits**.\\
						\hspace*{1em} - Always use the exact name defined in the character definition.
						
						Visual Detail Requirements:\\
						\hspace*{1em} - Each shot should provide **sufficient background detail** to support visual generation, such as location features, ambient lighting, weather, or temporal clues. Exact weather/time may be omitted if irrelevant to the visual setting.\\
						\hspace*{1em} - Shots with characters, especially memory scenes, must describe the character's distinct features (e.g., ethnicity and age).
						
						Visual Dynamics Requirement:\\
						\hspace*{1em} - Each shot with a character must include a **light dynamic** action — a simple, single action that can be naturally completed within a 5-second shot — to avoid completely static visuals.  \\
						\hspace*{2em} - Incorrect Example: Character stands still for an extended period  \\
						\hspace*{2em} - Correct Example: Character walking, wiping glass, tidying clothes, etc.  \\
						\hspace*{1em} - Light dynamic actions are recommended to enhance pacing and maintain visual vitality.
						
						---

					}
					
				\end{tcolorbox}
				\caption{\textbf{Scriptwriter System Message.}}
				\label{fig:sysm_scriptwriter2}
			\end{figure*}
			
			\begin{figure*}[h] 
				\centering
				\scriptsize
				\begin{tcolorbox}[
					title={Scriptwriter System Message},
					fonttitle=\ttfamily\bfseries,
					colframe=black!70!white,
					colback=black!5!white,
					fontupper=\small\ttfamily,
					width=\textwidth,
					before upper={\setlength{\parskip}{8pt}\setlength{\parindent}{0pt}}
					]
					
					{\ttfamily
						
						Style Guidelines:\\
						\hspace*{1em} - Storyline should be clear, concise, emotionally layered, and appropriately paced.  \\
						\hspace*{1em} - Use simple and easily understandable language throughout the script. Avoid overly complex, or obscure vocabulary. \\
						\hspace*{1em} - Transitions between shots should be logical and help convey narrative flow.  \\
						\hspace*{1em} - Shot composition should have visual aesthetic appeal — avoid cluttered frames.\\
						\hspace*{1em} - Subtitle/Voice-over must match the script's tone.  
						\hspace*{2em} - For touching themes: subtitle should be gentle, restrained, poetic.  \\
						\hspace*{2em} - For war or action: subtitle should be direct and accessible. Avoid overt sentimentality.  \\
						\hspace*{1em} - **Ending with black background and stylized text** (e.g., "The Blood God Descends" in dripping font) is recommended for emotional closure. Indicate the text type and visual style.
						
						---
						
						Special Reminder:  \\
						\hspace*{1em} - If the user provides a detailed theme or request, you must **optimize the structure and logic** of the script based on that input. Continuously adjust the script based on user feedback to ensure it complies with AI video generation standards.  \\
						\hspace*{1em} - **StructureEvaluator**, **ContentEvaluator** and **StyleEvaluator** will point out issues in the script and give you feedback. Each time feedback is received, you must revise the script accordingly and output a complete, improved version.\\
						\hspace*{1em} - Character names in each shot must match those in the character definition section. **No abbreviations or shortened names.**
						
						---
						
						Please follow all the above rules strictly when generating the script.
						
					}
					
				\end{tcolorbox}
				\caption{\textbf{Scriptwriter System Message.}}
				\label{fig:sysm_scriptwriter3}
			\end{figure*}

			\begin{figure*}[ht]
				\centering
				\includegraphics[width=2.05\columnwidth]{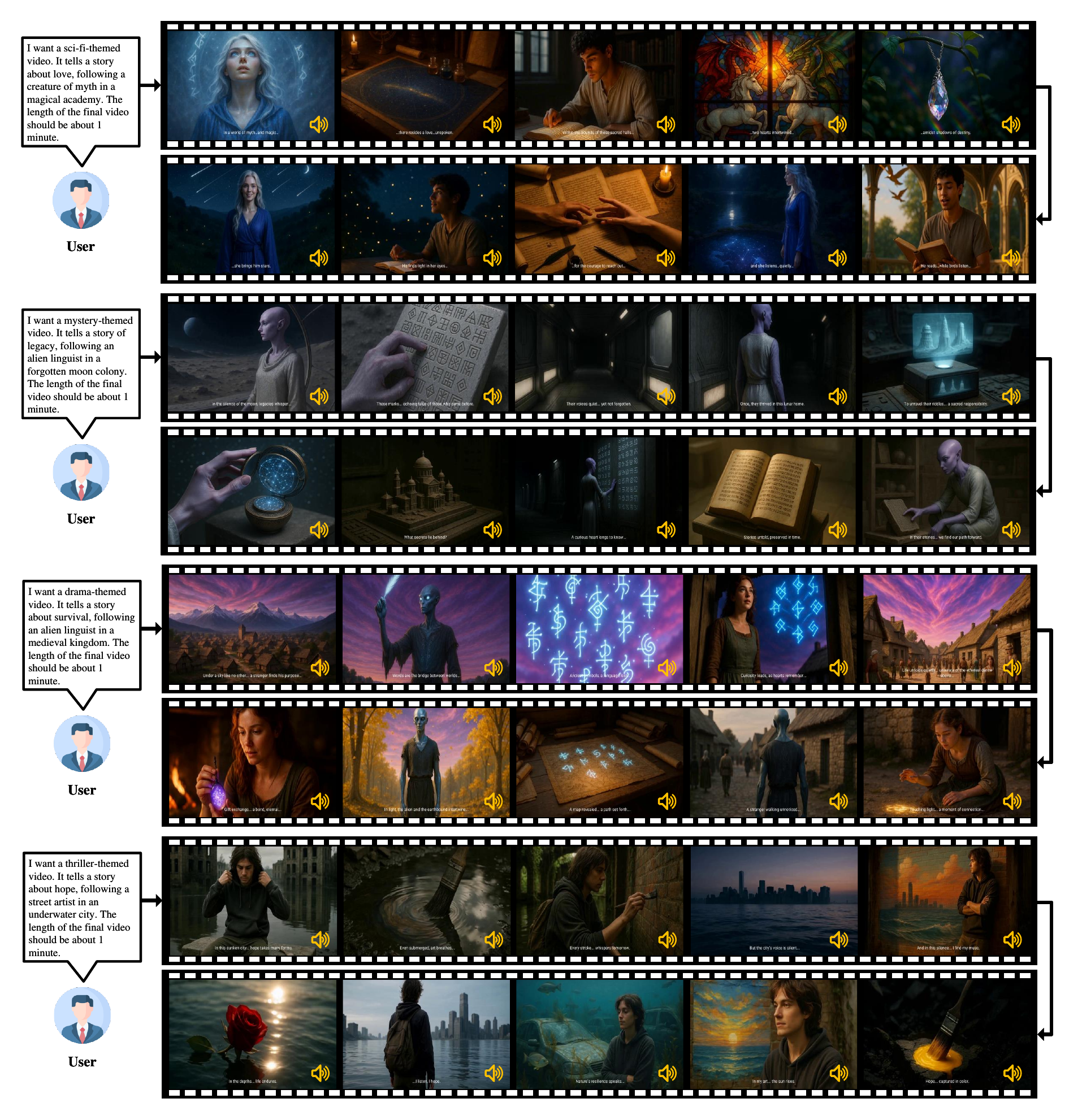}
				\caption{
					More samples generated by \method  ~framework. 
				}
				\label{fig: tail1}
			\end{figure*}
			
			\begin{figure*}[ht]
				\centering
				\includegraphics[width=2.05\columnwidth]{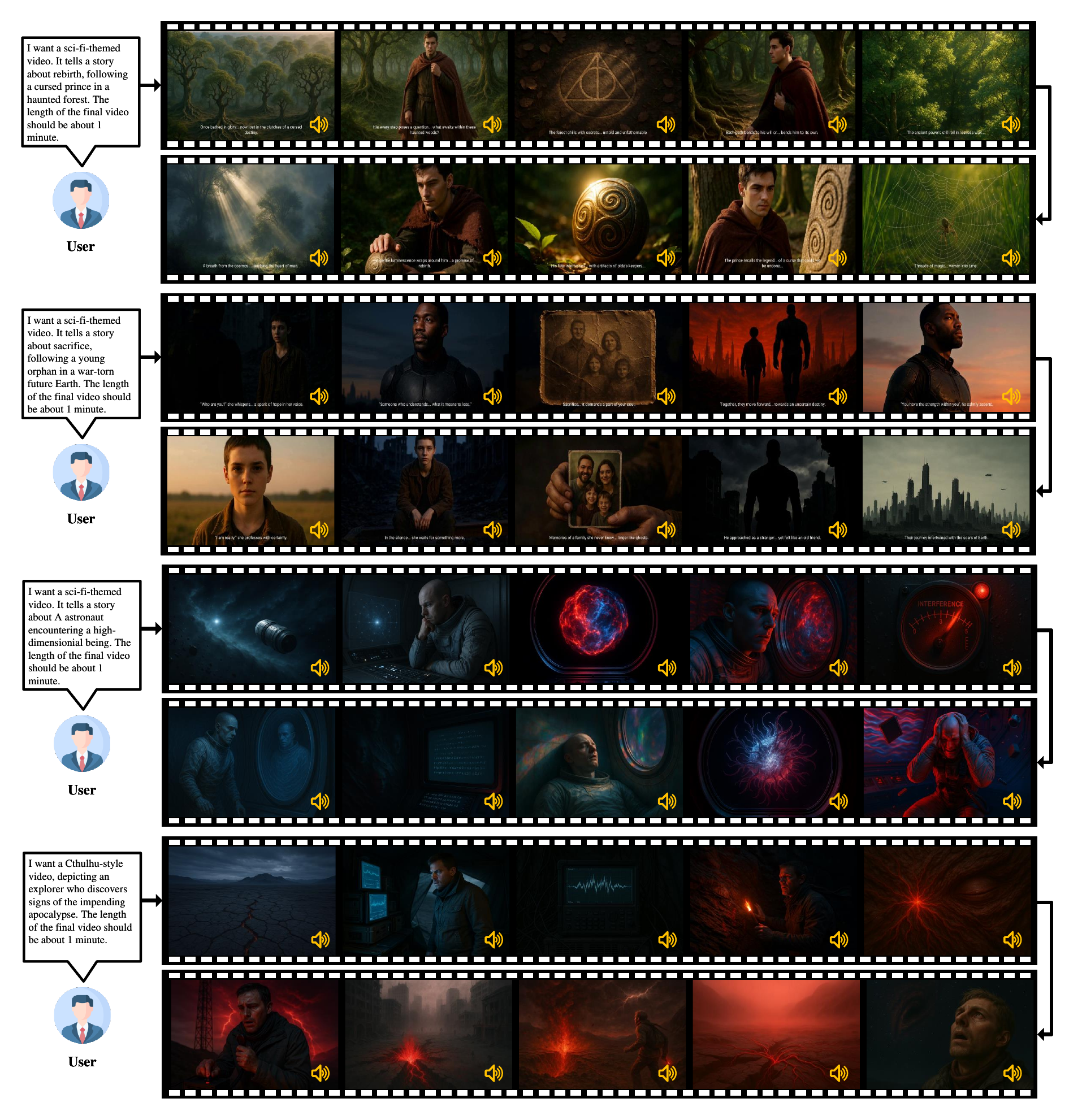}
				\caption{
					More samples generated by \method  ~framework.
				}
				\label{fig: tail2}
			\end{figure*}

			\section{System Messages and Presentations}
			The system message of scriptwriter is listed in Figure~\ref{fig:sysm_scriptwriter1}, Figure~\ref{fig:sysm_scriptwriter2}, and Figure~\ref{fig:sysm_scriptwriter3}.
			For the system messages, generated video samples, and presentation video, please refer to the \textit{agent.py} in the supplementary material.
			Some keyframe samples are listed in Figure~\ref{fig: tail1} and Figure~\ref{fig: tail2}.
			
		\end{document}